%% file: bare_jrnl.tex
\documentclass[journal]{IEEEtran}
\usepackage{tabularx}
\usepackage{multicol, multirow}
\usepackage{amssymb}
\usepackage{pifont}
\usepackage{graphicx}
\usepackage[dvipsnames]{xcolor}
\newcommand{\cmark}{\ding{51}}%
\newcommand{\xmark}{\ding{55}}%
\def\etal{{\em et al.\/}\,}
\usepackage[symbol]{footmisc}

\ifCLASSINFOpdf
\else
\fi
\begin{document}
%
\title{AI for IT Operations (AIOps) on Cloud Platforms: Reviews, Opportunities and Challenges}
%
%
%
\newcommand\blfootnote[1]{%
  \begingroup
  \renewcommand\thefootnote{}\footnote{#1}%
  \addtocounter{footnote}{-1}%
  \endgroup
}

\author{%
  \IEEEauthorblockN{%
    Qian Cheng \IEEEauthorrefmark{1}\IEEEauthorrefmark{2},  
    Doyen Sahoo \IEEEauthorrefmark{1}, 
    Amrita Saha,
    Wenzhuo Yang, 
    Chenghao Liu,
    Gerald Woo,\\
    Manpreet Singh,
    Silvio Saverese, and
    Steven C. H. Hoi\\
  }%
\IEEEauthorblockA{ Salesforce AI}%
}

\maketitle

\begin{abstract}
Artificial Intelligence for IT operations (AIOps) aims to combine the power of AI with the big data generated by IT Operations processes, particularly in cloud infrastructures, to provide actionable insights with the primary goal of maximizing availability. There are a wide variety of problems to address, and multiple use-cases, where AI capabilities can be leveraged to enhance operational efficiency. Here we provide a review of the AIOps vision, trends challenges and opportunities, specifically focusing on the underlying AI techniques. We discuss in depth the key types of data emitted by IT Operations activities, the scale and challenges in analyzing them, and where they can  be helpful. We categorize the key AIOps tasks as - incident detection, failure prediction, root cause analysis and automated actions. We discuss the problem formulation for each task, and then present a taxonomy of techniques to solve these problems. We also identify relatively under explored topics, especially those that could significantly benefit from advances in AI literature. We also provide insights into the trends in this field, and what are the key investment opportunities.
\end{abstract}

\blfootnote{ \IEEEauthorrefmark{1} Equal Contribution}
\blfootnote{ \IEEEauthorrefmark{2} Work done when author was with Salesforce AI}


%
\IEEEpeerreviewmaketitle

\section{Introduction}\label{sec:intro}

\input{intro}

%
%
%
%

\section{Contribution of This Survey}\label{sec:overview}
\input{survey_overview}

\section{Data for AIOps}\label{sec:problems}
\input{observability}

\section{Incident Detection}
\label{sec:incident_detection}
\input{incident_detection}

\section{Failure Prediction}
\label{sec:failure_prediction}
\input{failure_prediction}

\section{Root Cause Analysis}
\label{sec:rca}
\input{root_cause_analysis}

\section{Automated Actions}
\input{automated_actions.tex}




\section{Future of AIOps}
\label{sec:future}
\input{future}

\section{Conclusion}
\label{sec:conclusion}
\input{conclusion}


%


\section*{Acknowledgment}
We want to thank all participants who took the time to accomplish this survey. Their knowledge and experiences about AI fundamentals were invaluable to our study. We are also grateful to our colleagues at the Salesforce AI Research Lab and collaborators from other organizations for their helpful feedback and support.

\appendices
\section{Terminology}
\input{appendix_terminology.tex}


\onecolumn
\section{Tables}
\input{appendix_tables.tex}

\twocolumn
\ifCLASSOPTIONcaptionsoff
  \newpage
\fi



%


\bibliographystyle{IEEEtran}
\bibliography{bibtex/bib/log,bibtex/bib/log_survey,bibtex/bib/qcheng_aiops,bibtex/bib/metric_ad,bibtex/bib/wenzhuo}

%








\end{document}

%% file: intro.tex
Modern software has been evolving rapidly during the era of digital transformation. New infrastructure, techniques and design patterns - such as cloud computing, Software-as-a-Service (SaaS), microservices, DevOps, etc. have been developed to boost software development. Managing and operating the infrastructure of such modern software is now facing new challenges. For example, when traditional software transits to SaaS, instead of handing over the installation package to the user, the software company now needs to provide 24/7 software access to all the subscription based users. Besides developing and testing, service management and operations are now the new set of duties of SaaS companies. Meanwhile, traditional software development separates functionalities of the entire software lifecycle. Coding, testing, deployment and operations are usually owned by different groups. Each of these groups requires different sets of skills. However, agile development and DevOps start to obfuscate the boundaries between each process and DevOps engineers are required to take E2E responsibilities. Balancing development and operations for a DevOps team become critical to the whole team's productivity. 

Software services need to guarantee service level agreements (SLAs) to the customers, and often set internal Service Level Objectives (SLOs). Meeting SLAs and SLOs is one of the top priority for CIOs to choose the right service providers\cite{Olavsrud2012}. Unexpected service downtime can impact availability goals and cause significant financial and trust issues. For example, AWS experienced a major service outage in December 2021, causing multiple first and third party websites and heavily used services to experience downtime \cite{aws2021}. 

IT Operations plays a key role in the success of modern software companies and as a result multiple concepts have been introduced, such as IT service management (ITSM) specifically for SaaS, and IT operations management (ITOM) for general IT infrastructure. These concepts focus on different aspects IT operations but the underlying workflow is very similar. Life cycle of Software systems can be separated into several main stages, including planning, development/coding, building, testing, deployment, maintenance/operations, monitoring, etc. \cite{Olavsrud2021}. The operation part of DevOps can be further broken down into four major stages: observe, detect, engage and act, shown in Figure \ref{fig:ops_lifecycle}. Observing stage includes tasks like collecting different telemetry data (metrics, logs, traces, etc.), indexing and querying and visualizing the collected telemetries. Time-to-observe (TTO) is a metric to measure the performance of the observing stage. Detection stage includes tasks like detecting incidents, predicting failures, finding correlated events, etc. whose performance is typically measured as the Time-to-detect (TTD) (in addition to precision/recall). Engaging stage includes tasks like issue triaging, localization, root-cause analysis, etc., and the performance is often measured by Time-to-triage (TTT). Acting stage includes immediate remediation actions such as reboot the server, scale-up / scale-out resources, rollback to previous versions, etc. Time-to-resolve (TTR) is the key metric measured for the acting stage. Unlike software development and release, where we have comparatively mature continuous integration and continuous delivery (CI/CD) pipelines, many of the post-release operations are often done manually. Such manual operational processes face several challenges:

\begin{figure}[ht]
    \centering
    \includegraphics[width=0.49\textwidth]{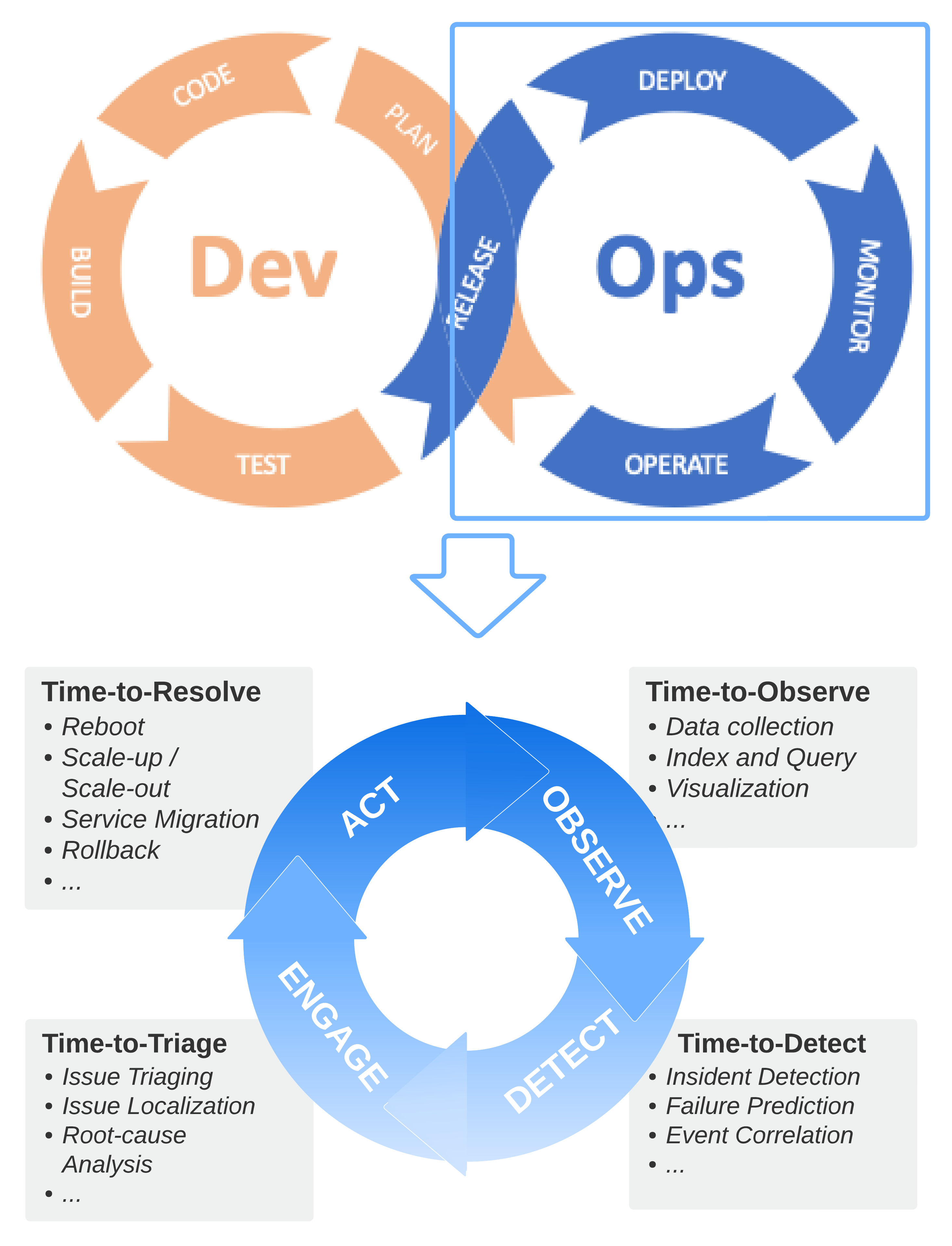}
    \caption{Common DevOps life cycles\cite{Olavsrud2021} and ops breakdown. Ops can comprise four stages: observe, detect, engage and act. Each of the stages has a corresponding measure: time-to-observe, time-to-detect, time-to-triage and time-to-resolve.}
    \label{fig:ops_lifecycle}
\end{figure}

\begin{itemize}
    \item \textit{Manual operations struggle to scale.} The capacity of manual operations is limited by the size of the DevOps team and the team size can only increase linearly. When the software usage is at growing stage, the throughput and workloads may grow exponentially, both in scale and complexity. It is difficult for DevOps team to grow at the same pace to handle the increasing amount of operational workload. 
    \item \textit{Manual operations is hard to standardize.} It is very hard to keep the same high standard across the entire DevOps team given the diversity of team members  (e.g. skill level, familiarity with the service, tenure, etc.).  It takes significant amount of time and effort to grow an operational domain expert who can effectively handle incidents. Unexpected attrition of these experts could significantly hurt the operational efficiency of a DevOps team. 
    \item \textit{Manual operations are error-prone.} It is very common that human operation error causes major incidents. Even for the most reliable cloud service providers, major incidents have been caused by human error in recent years.
\end{itemize}

Given these challenges, fully-automated operations pipelines powered by AI capabilities becomes a promising approach to achieve the SLA and SLO goals. AIOps, an acronym of AI for IT Operations, was coined by Gartner at 2016. According to Gartner Glossary, "AIOps combines big data and machine learning to automate IT operations processes, including event correlation, anomaly detection and causality determination"\cite{gartner}. In order to achieve fully-automated IT Operations, investment in AIOps technolgies is imperative. AIOps is the key to achieve \textit{high availability, scalability and operational efficiency}. For example, AIOps can use AI models can automatically analyze large volumes of telemetry data to detect and diagnose incidents much faster, and much more consistently than humans, which can help achieve ambitious targets such as 99.99 availability. AIOps can dynamically scale its capabilities with growth demands and use AI for automated incident and resource management, thereby reducing the burden of hiring and training domain experts to meet growth requirements. Moreover, automation through AIOps helps save valuable developer time, and avoid fatigue. AIOps, as an emerging AI technology, appeared on the trending chart of Gartner Hyper Cycle for Artificial Intelligence in 2017 \cite{Siddique2018}, along with other popular topics such as deep reinforcement learning, nature-language generation and artificial general intelligence. As of 2022, enterprise AIOps solutions have witnessed increased adoption by many companies' IT infrastructure. The AIOps market size is predicted to be \$11.02B by end of 2023 with cumulative annual growth rate (CAGR) of 34\%.

\begin{figure}[ht]
    \centering
    \includegraphics[width=0.49\textwidth]{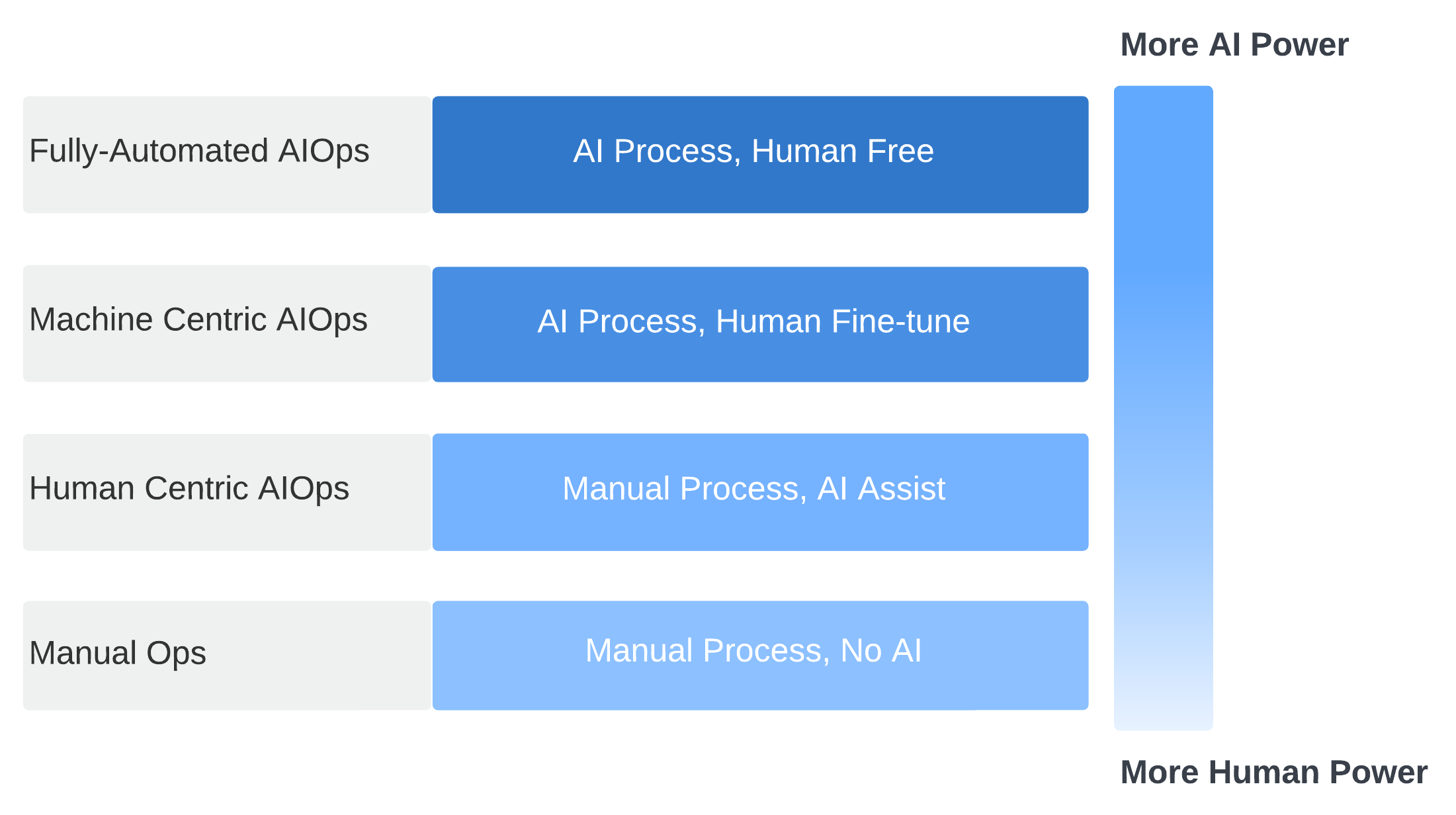}
    \caption{AIOps Transformation. Different maturity levels based on adoption of AI techniques: Manual Ops, human-centric AIOps, machine-centric AIOps, fully-automated AIOps.}
    \label{fig:aiops_levels}
\end{figure}

AIOps comprises a set of complex problems. Transforming from manual to automated operations using AIOps is not a one-step effort. Based on the adoption level of AI techniques, we break down AIOps maturity into four different levels based on the adoption of AIOps capabilities as shown in Figure \ref{fig:aiops_levels}.

\textbf{Manual Ops.} At this maturity level, DevOps follows traditional best practices and all processes are setup manually. There is no AI or ML models. This is the baseline to compare with in AIOps transformation.

\textbf{Human-centric}. At this level, operations are done mainly in manual process and AI techniques are adopted to replace sub-procedures in the workflow, and mainly act as assistants. For example, instead of glass watching for incident alerts, DevOps or SREs can set dynamic alerting threshold based on anomaly detection models. Similarly, the root cause analysis process requires watching multiple dashboards to draw insights, and AI can help automatically obtain those insights.

\textbf{Machine-centric}. At this level, all major components (monitoring, detecting, engaging and acting) of the E2E operation process are empowered by more complex AI techniques. Humans are mostly hands-free but need to participate in the human-in-the-loop process to help fine-tune and improve the AI systems performance. For example, DevOps / SREs operate and manage the AI platform to guarantee training and inference pipelines functioning well, and domain experts need to provide feedback or labels for AI-made decisions to improve performance. 

\textbf{Fully-automated}. At this level, AIOps platform achieves full automation with minimum or zero human intervention. With the help of fully-automated AIOps platforms, the current CI/CD (continuous integration and continuous deployment) pipelines can be further extended to CI/CD/CM/CC (continuous integration, continuous deployment, continuous monitoring and continuous correction) pipelines.

Different software systems, and companies may be at different levels of AIOps maturity, and their priorities and goals may differ with regard to specific AIOps capabilities to be adopted. Setting up the right goals is important for the success of AIOps applications. We foresee the trend of shifting from manual operation all the way to fully-automated AIOps in the future, with more and more complex AI techniques being used to address challenging problems. In order to enable the community to adopt AIOps capabilities faster, in this paper, we present a comprehensive survey on the various AIOps problems and tasks and the solutions developed by the community to address them.


%% file: survey_overview.tex
Increasing number of research studies and industrial products in the AIOps domain have recently emerged to address a variety of problems. Sabharwal \etal  published a book "Hands-on AIOps" to discuss practical AIOps and implementation \cite{Sabharwal}. Several AIOps literature reviews are also accessible \cite{Dang2019} \cite{Rijal2022} to help audiences better understand this domain. However, there are very limited efforts to provide a holistic view to deeply connect AIOps with latest AI techniques. Most of the AI related literature reviews are still topic-based, such as deep learning anomaly detection \cite{chalapathy2019deep} \cite{Akoglu2015}, failure management, root-cause analysis \cite{Soldani2021}, etc. There is still limited effort to provide a holistic view about AIOps, covering the status in both academia and industry. We prepare this survey to address this gap, and focus more on AI techniques used in AIOps. 

\begin{figure*}[ht]
    \centering
    \includegraphics[width=\textwidth]{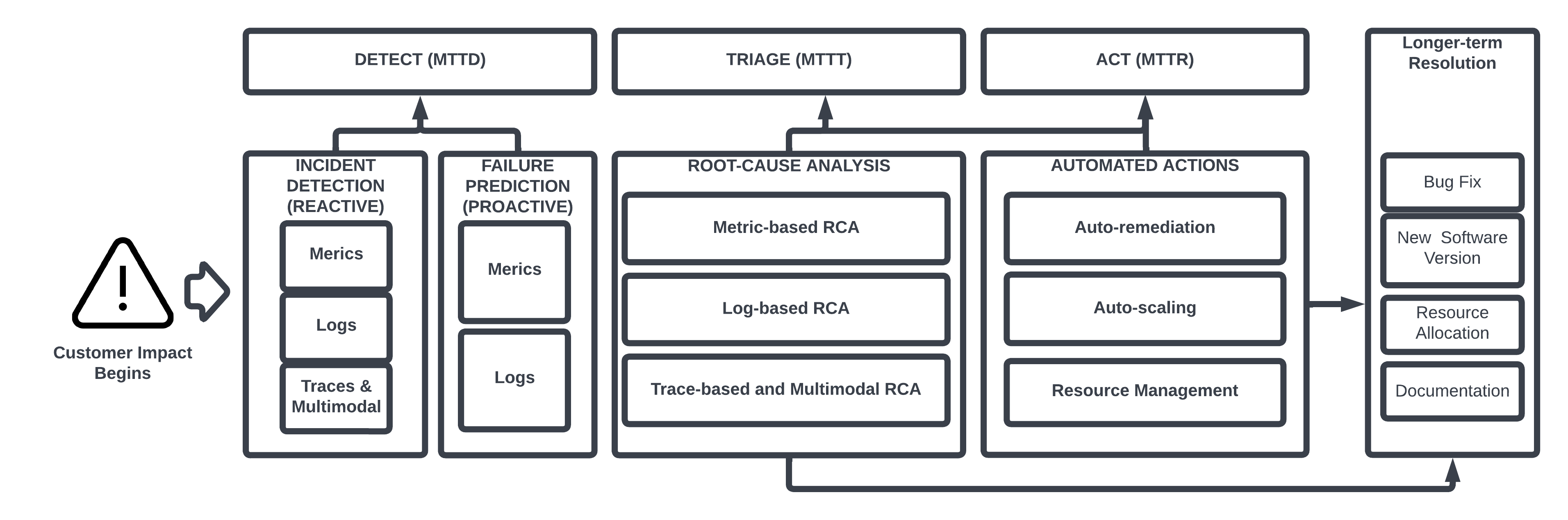}
    \caption{AIOps Tasks. In this survey we discuss a series of AIOps tasks, categorized by which operational stages these tasks contribute to, and the observability data type it takes.}
    \label{fig:aiops_tasks}
\end{figure*}

Except for the monitoring stage, where most of the tasks focus on telemetry data collection and management, AIOps covers the other three stages where the tasks focus more on analytics. In our survey, we group AIOps tasks based on which operational stage they can contribute to, shown in Figure \ref{fig:aiops_tasks}.

\textbf{Incident Detection.} Incident detection tasks contribute to detection stage. The goal of these tasks are reducing mean-time-to-detect (MTTD). In our survey we cover time series incident detection (Section \ref{sec:incident_metric}), log incident detection (Section \ref{sec:id_logs}), trace and multimodal incident detection (Section \ref{sec:trace_ad}).

\textbf{Failure Prediction.} Failure prediction also contributes to detection stage. The goal of failure prediction is to predict the potential issue before it actually happens so actions can be taken in advance to minimize impact. Failure prediction also contributes to reducing mean-time-to-detect (MTTD). In our survey we cover metric failure prediction (Section \ref{sec:failure_metrics}) and log failure prediction (Section \ref{sec:failure_logs}). There are very limited efforts in literature that perform traces and multimodal failure prediction. 

\textbf{Root-cause Analysis.} Root-cause analysis tasks contributes to multiple operational stages, including triaging, acting and even support more efficient long-term issue fixing and resolution. Helping as an immediate response to an incident, the goal is to minimize time to triage (MTTT), and simultaneously contribute to reduction on reducing Mean Time to Resolve (MTTR). An added benefit is also reduction in human toil. We further breakdown root-cause analysis into time-series RCA (Section \ref{sec:rca_logs}), logs RCA (Section \ref{sec:rca_logs}) and traces and multimodal RCA (Section \ref{sec:trace_rca}).

\textbf{Automated Actions.} Automated actions contribute to acting stage, where the main goal is to reduce mean-time-to-resolve (MTTR), as well as long-term issue fix and resolution. In this survey we discuss about a series of methods for auto-remediation (Section \ref{sec:auto_remediation}), auto-scaling (Section \ref{sec:auto-scaling}) and resource management (Section \ref{sec:resource_management}).

%% file: observability.tex



Before we dive into the problem settings, it is important to understand the data available to perform AIOps tasks. Modern software systems generate tremendously large volumes of observability metrics. The data volume keeps growing exponentially with digital transformation \cite{Davidovski2018}. The increase in the volume of data stored in large unstructured Data lake systems makes it very difficult for DevOps teams to consume the new information and fix consumers’ problems efficiently \cite{Battina2021}. Successful products and platforms are now built to address the monitoring and logging problems. Observability platforms, e.g. Splunk, AWS Cloudwatch, are now supporting emitting, storing and querying large scale telemetry data. 

Similar to other AI domains, observability data is critical to AIOps. Unfortunately there are limited public datasets in this domain and many successful AIOps research efforts are done with self-owned production data, which usually are not available publicly.  In this section, we describe major telemetry data type including metrics, logs, traces and other records, and present a collection of public datasets for each data type.

\subsection{Metrics}
\input{metric_observability}

\subsection{Logs}

\input{log_observability}

\subsection{Traces} 

\begin{figure}
    \centering
    \includegraphics[width=0.4\textwidth]{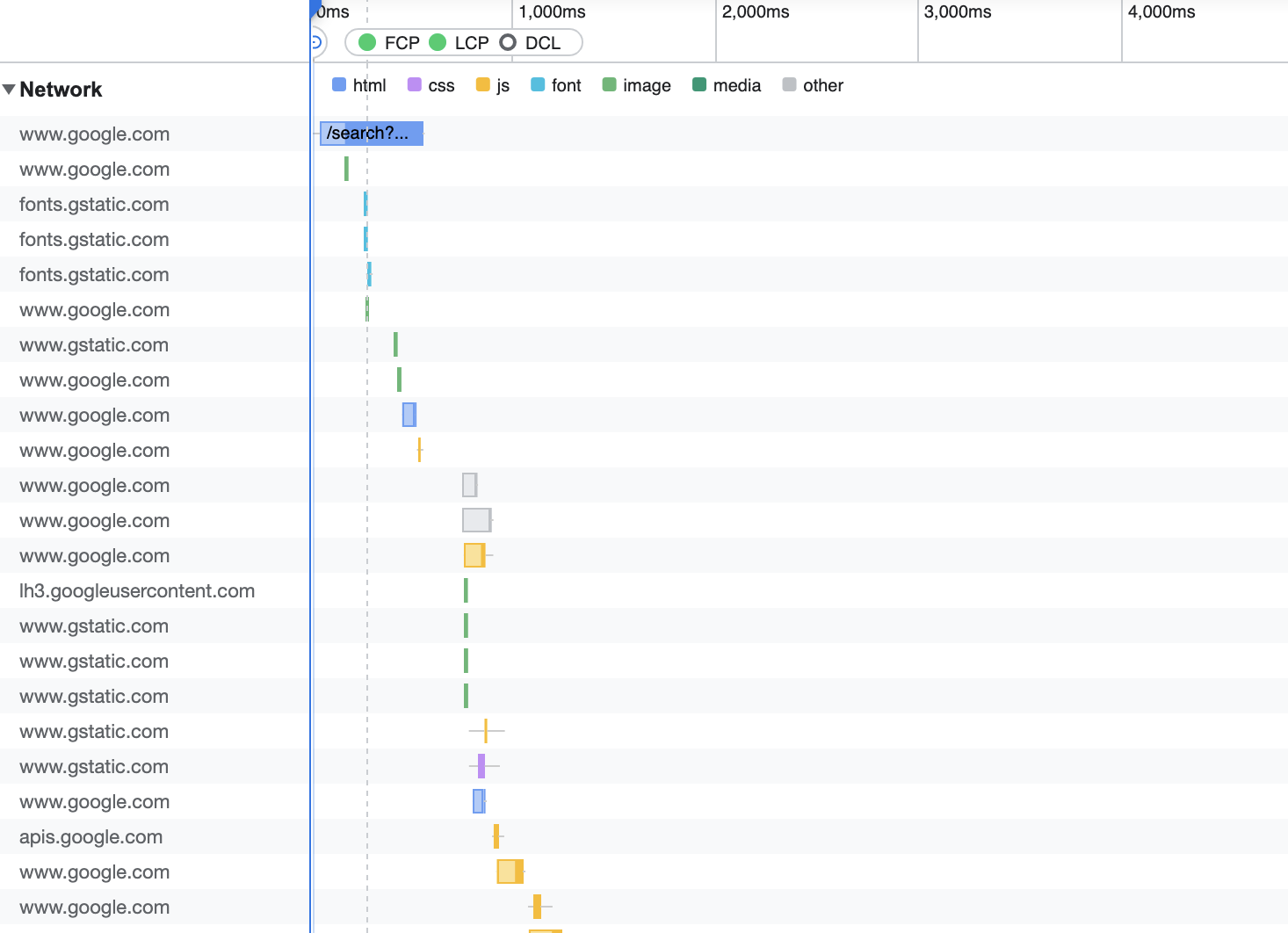}
    \caption{An snapshot of trace graph of user requests when using Google Search.}
    \label{fig:trace_graph}
\end{figure}

Trace data are usually presented as semi-structured logs, with identifiers to reconstruct the topological maps of the applications and network flows of target requests. For example, when user uses Google search, a typical trace graph of this user request looks like in Figure \ref{fig:trace_graph}. Traces are composed system events (spans) that tracks the entire progress of a request or execution. A span is a sequence of semi-structured event logs. Tracing data makes it possible to put different data modality into the same context. Requests travel through multiple services / applications and each application may have totally different behavior. Trace records usually contains two required parts: timestamps and span\_id. By using the timestamps and span\_id, we can easily reconstruct the trace graph from trace logs. 

Trace analysis requires reliable tracing systems. Trace collection systems such as ReTrace \cite{Sheldon2007} can help achieve fast and inexpensive trace collections. Trace collectors are usually code agnostic and can emit different levels of performance trace data back to the trace stores in near real-time. Early summarization is also involved in the trace collection process to help generate fine-grained events \cite{Fonseca2007}.

Although trace collection is common for system observability, it is still challenging to acquire high quality trace data to train AI models. As far as we know, there are very few public trace datasets with high quality labels. Also the only few existing public trace datasets like \cite{Zhou2014} are not widely adopted in AIOps research. Instead, most AIOps related trace analysis research use self-owned production or simulation trace data, which are generally not available publicly. 

\subsection{Other Data}

Besides the machine generated observability data like metrics, logs, traces, etc., there are other types of operational data that could be used in AIOps. 

Human activity records is part of these valuable data. Ticketing systems are used for DevOps/SREs to communicate and efficiently resolve the issues. This process generates large amount of human activity records. The human activity data contains rich knowledge and learnings about solutions to existing issues, which can be used to resolve similar issues in the future. 

User feedback data is also very important to improve AIOps system performance. Unlike the issue tickets where human needs to put lots of context information to describe and discuss the issue, user feedback can be as simple as one click to confirm if the alert is good or bad. Collecting real-time user feedback of a running system and designing human-in-the-loop workflows are also very significant for success of AIOps solutions.

Although many companies collects these types of data and use them to improve their operation workflows, there are still very limited published research discussing how to systematically incorporate these other types of operational data in AIOps solutions. This brings challenges as well as opportunities to make further improvements in AIOps domain.

Next, we discuss the key AIOps Tasks - Incident Detection, Failure Prediction, Root Cause Analysis, and Automated Actions, and systematically review the key contributions in literature in these areas.

%% file: metric_observability.tex
Metrics are numerical data measured over time which provide a snapshot of the system behavior. 
Metrics can represent a broad range of information, broadly classified into compute metrics and service metrics. Compute metrics (e.g. CPU utilization, memory usage, disk I/O) are an indicator of the health status of compute nodes (servers, virtual machines, pods). They are collected at the system level using tools such as Slurm \cite{yoo2003slurm} for usage statistics from jobs and nodes, and the Lustre parallel distributed file system for I/O information. Service metrics (e.g. request count, page visits, number of errors) measure the quality and level of service of customer facing applications. Aggregate statistics of such numerical data also fall under the category of metrics, providing a more coarse-grained view of system behavior.

Metrics are constantly generated by all components of the cloud platform life cycle, making it one of the most ubiquitous forms of AIOps data. 
Cloud platforms and supercomputer clusters can generate petabytes of metrics data, making it a challenge to store and analyze, but at the same time, brings immense observability to the health of the entire IT operation. 
Being numerical time-series data, metrics are simple to interpret and easy to analyze, allowing for simple threshold-based rules to be acted upon. 
At the same time, they contain sufficiently rich information to be used to power more complex AI based alerting and actions. 

The major challenge in leveraging insights from metrics data arises due to their diverse nature.
Metrics data can exhibit a variety of patterns, such as cyclical patterns (repeating patterns hourly, daily, weekly, etc.), sparse and intermittent spikes, and noisy signals. The characteristics of the metrics ultimately depend on the underlying service or job.

In Table \ref{tab:metrics_observability}, we briefly describe the datasets and benchmarks of metrics data. Metrics data have been used in studies characterizing the workloads of cloud data centers, as well as the various AIOps tasks of incident detection, root cause analysis, failure prediction, and various planning and optimization tasks like auto-scaling and VM pre-provisioning.

\begin{figure}
    \centering
    \includegraphics[width=\columnwidth]{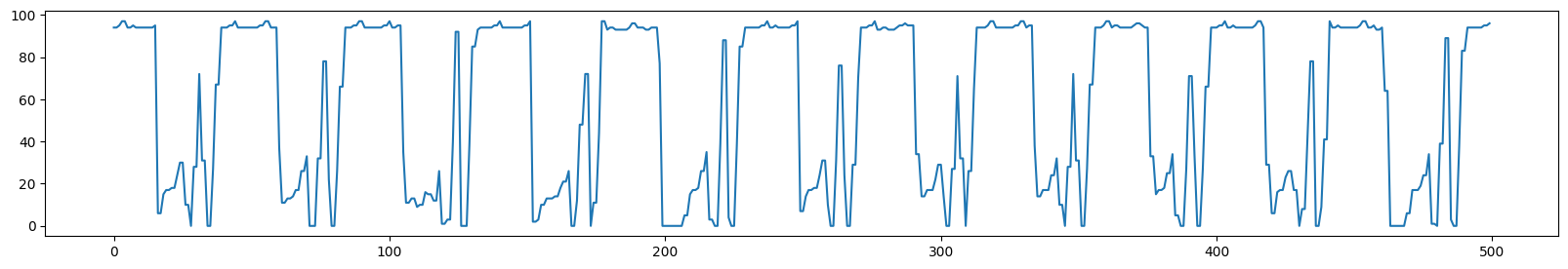}
    \includegraphics[width=\columnwidth]{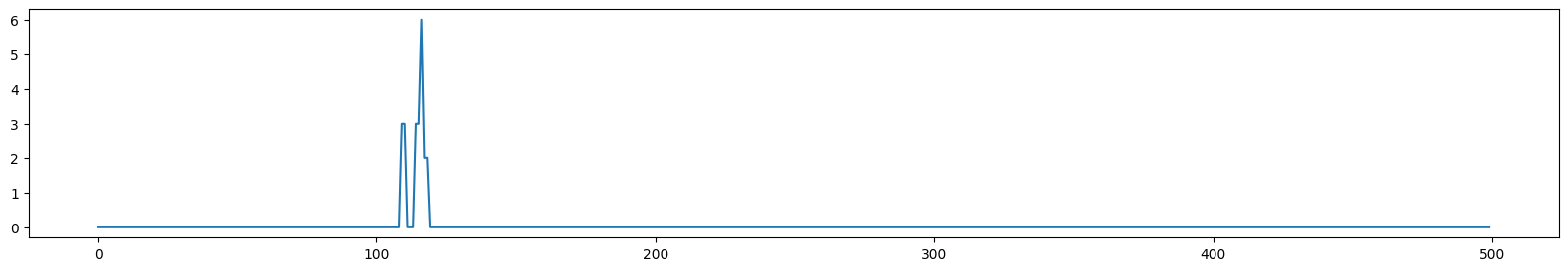}
    \includegraphics[width=\columnwidth]{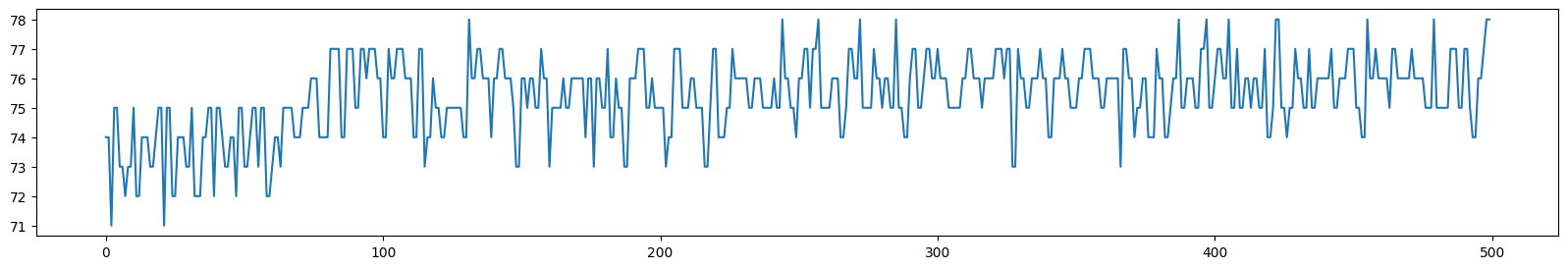}
    \caption{GPU utilization metrics from the MIT Supercloud Dataset exhibiting various patterns (cyclical, sparse and intermittant, noisy).}
    \label{fig:metrics_example}
\end{figure}





%% file: log_observability.tex
Software logs are specifically designed by the software developers in order to record any type of runtime information about processes executing within a system - thus making them an ubiquitous part of any modern system or software maintenance. Once the system is live and throughout its life-cycle, it continuously emits huge volumes of such logging data which naturally contain a lot of rich  dynamic runtime information relevant to IT Operations and Incident Management of the system. Consequently in AI driven IT-Ops pipelines, automated log based analysis plays an important role in Incident Management - specifically in tasks like Incident Detection and Causation and Failure Prediction, as have been studied by multiple literature surveys in the past \cite{10.1007/s11036-021-01832-3,10.1145/3460345,10.1145/3483424,10.1145/3459637.3482209,10.1145/3510003.3510155,10.1145/3468264.3473933,DBLP:journals/corr/abs-2112-03159,10.1145/3501297,9716129}. 

In most of the practical cases, especially in industrial settings, the volume of the logs can go upto an order of petabytes of loglines per week. Also because of the nature of log content, log data dumps are much more heavier in size in comparison to time series telemetry data. This requires special handling of logs observability data in form of data streams, - where today, there are various services like Splunk, Datadog, LogStash, NewRelic, Loggly, Logz.io etc employed to efficiently store and access the log stream and also visualize, analyze and query past log data using specialized structured query language. 

\vspace{1em}
\textbf{Nature of Log Data.} Typically these logs consist of semi-structured data i.e. a combination of structured and unstructured data. Amongst the typical types of unstructured data there can be natural language tokens, programming language constructs (e.g. method names) and the structured part can consist of quantitative or categorical telemetry or observability metrics data, which are printed in runtime by various logging statements embedded in the source-code or sometimes generated automatically via loggers or logging agents. Depending on the kind of service the logs are dumped from, there can be a diverse types of logging data with heterogeneous form and content. For example, logs can be originating from distributed systems (e.g. hadoop or spark), operating systems (windows or linux) or in complex supercomputer systems or can be dumped at hardware level (e.g. switch logs) or middle-ware level (like servers e.g. Apache logs) or by specific applications (e.g. Health App). Typically each logline comprises of a fixed part which is the template that had been designed by the developer and some variable part or parameters which capture some runtime information about the system. 

\vspace{1em}
\textbf{Complexities of Log Data.} Thus, apart from being one of the most generic and hence crucial data-sources in IT Ops, logs are one of the most complex forms of observability data due to their open-ended form and level of granularity at which they contain system runtime information. In cloud computing context, logs are the source of truth for cloud users to the underlying servers that running their applications since cloud providers don't grant full access to their users of the servers and platforms. Also, being designed by developers, logs are immediately affected by any changes in the source-code or logging statements by developers. This results in non-stationarity in the logging vocabulary or even the entire structure or template underlying the logs.

\vspace{1em}
\textbf{Log Observability Tasks.} Log observability typically involves different tasks like anomaly detection over logs during incident detection (Section \ref{sec:id_logs}), root cause analysis over logs (Section \ref{sec:rca_logs}) and log based failure prediction (Section \ref{sec:failure_logs}). 

\vspace{1em}
\textbf{Datasets and Benchmarks.} Out of the different log observability tasks, log based anomaly detection is one of the most objective tasks and hence most of the publicly released benchmark datasets have been designed around anomaly detection. In Table \ref{tab:log_datasets}, we give a comprehensive description about the different public benchmark datasets that have been used in the literature for anomaly detection tasks. Out of these, datasets Switch and subsets of HPC and BGL have also been redesigned to serve failure prediction task. On the other hand there are no public benchmarks on log based RCA tasks, which has been typically evaluated on private enterprise data.

\begin{figure}
    \centering
    \includegraphics[width=\columnwidth]{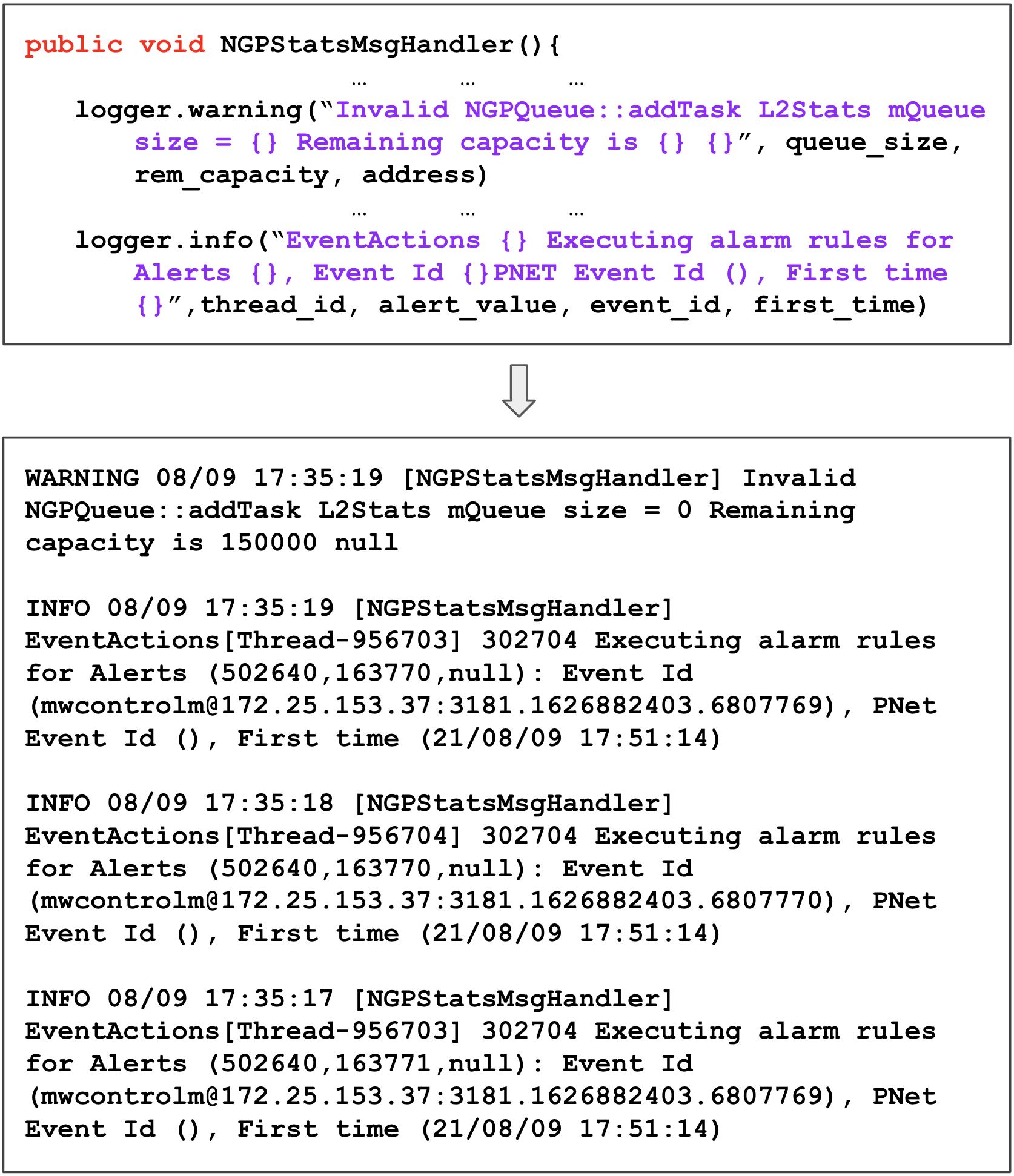}
    \caption{An example of Log Data generated in IT Operations}
    \label{fig:log_example}
\end{figure}

%% file: incident_detection.tex
Incident detection employs a variety of anomaly detection techniques. Anomaly detection is to detect abnormalities, outliers or generally events that not normal. In AIOps context, anomaly detection is widely adopted in detecting any types of abnormal system behaviors. To detect such anomalies, the detectors need to utilize different telemetry data, such as metrics, logs, traces. Thus, anomaly detection can be further broken down to handling one or more specific telemetry data sources, including metric anomaly detection, log anomaly detection, trace anomaly detection. Moreover, multi-modal anomaly detection techniques can be employed if multiple telemetry data sources are involved in the detection process. In recent years, deep learning based anomaly detection techniques \cite{chalapathy2019deep} are also widely discussed and can be utilized for anomaly detection in AIOps.  Another way to distinguish anomaly detection techniques is depending on different application use cases, such as detecting service health issues, detecting networking issues, detecting security issues, fraud transactions, etc. Usually these variety of techniques are derived from same set of base detection algorithms and localized to handle specific tasks. From technical perspective, detecting anomalies from different telemetry data sources are better aligned with the AI technology definitions, such as, metric are usually time-series, logs are text / natural language, traces are event sequences/graphs, etc. In this article,  we discuss anomaly detection by different telemetry data sources.

\subsection{Metrics based Incident Detection}
\label{sec:incident_metric}

\input{metric_ad}

\subsection{Logs based Incident Detection}
\label{sec:id_logs}
\input{log_ad}

\subsection{Traces and Multimodal Incident Detection}
\label{sec:trace_ad}
\input{trace_ad}

%% file: metric_ad.tex
\vspace{1em}
\textbf{Problem Definition} 

To ensure the reliability of services, billions of metrics are constantly monitored and collected at equal-space timestamp \cite{DBLP:conf/issre/ZhangZSSSZPW21}. Therefore, it is straightforward to organize metrics as time series data for subsequent analysis. Metric based incident detection, which aims to find the anomalous behaviors of monitored metrics that significantly deviate from the other observations, is vital for operators to timely detect software failures and trigger failure diagnosis to mitigate loss. The most basic form of incident detection on metrics is the rule-based method which sets up an alert when a metric breaches a certain threshold. Such an approach is only able to capture incidents which are defined by the metric exceeding the threshold, and is unable to detect more complex incidents. 
The rule-based method to detect incidents on metrics are generally too naive, and only able to account for the most simple of incidents. They are also sensitive to the threshold, producing too many false positives when the threshold is too low, and false negatives when the threshold is too high. Due to the open-ended nature of incidents, increasingly complex architectures of systems, and increasing size of these systems and number of metrics, manual monitoring and rule-based methods are no longer sufficient. Thus, more advanced metric-based incident detection methods that leveraging AI capability is urgent. 

As metrics are a form of time series data, and incidents are expressed as an abnormal occurrence in the data, metric incident detection is most often formulated as a time series anomaly detection problem \cite{braei2020anomaly, blazquez2021review, choi2021deep}. 
In the following, we focus on the AIOps setting and categorize it based on several key criteria: (i) learning paradigm, (ii) dimensionality, (iii) system, and (iv) streaming updates. We further summarize a list of time series anomaly detection methods with a comparison over these criteria in Table \ref{tab:metric_ad}.

\vspace{1em}
\textbf{Learning Setting}
\paragraph{Label Accessibility} One natural way to formulate the anomaly detection problem, is as the supervised binary classification problem, to classify whether a given observation is an anomaly or not \cite{liu2015opprentice, gao2020robusttad}. Formulating it as such has the benefit of being able to apply any supervised learning method, which has been intensely studied in the past decades \cite{han2022adbench}. However, due to the difficulty in obtaining labelled data for metrics incident detection \cite{li2022constructing} and labels of anomalies are prone to error \cite{DBLP:journals/corr/abs-2009-13807}, unsupervised approaches, which do not require labels to build anomaly detectors, are generally preferred and more widespread. Particularly, unsupervised anomaly detection methods can be roughly categorized into density-based methods, clustering-based methods, and reconstruction-based methods \cite{braei2020anomaly, blazquez2021review, choi2021deep}. Density-based methods compute local density and local connectivity for outlier decision. Clustering-based methods formulate the anomaly score as the distance to cluster center. Reconstruction-based methods explicitly model the generative process of the data and measure the anomaly score with the reconstruction error. While methods in metric anomaly detection are generally unsupervised, there are cases where there is some access to labels. In such situations, semi-supervised, domain adaptation, and active learning paradigms come into play. 
The semi-supervised paradigm \cite{xu2018unsupervised, bu2018rapid, huang2022semi} enables unsupervised models to leverage information from sparsely available positive labels \cite{li2005learning}.
Domain adaptation \cite{zhang2019cross} relies on a labelled source dataset, while the target dataset is unlabeled, with the goal of transferring a model trained on the source dataset, to perform anomaly detection on the target. 

\paragraph{Streaming Update} Since metrics are collected in large volume every minute, the model is used online to detect anomalies. It is very common that temporal patterns of metrics change overtime. The ability to perform timely model updates when receiving new incoming data is an important criteria. On the one hand, conventional models can handle the data stream via retraining the whole model periodically \cite{liu2015opprentice,laptev2015generic,gao2020robusttad, huang2022semi}. However, this strategy could be computationally expensive, and bring extra non-trivial questions, such as, how often should this retraining be performed. On the other hand, some methods \cite{guha2016robust, ahmad2017unsupervised} have efficient updating mechanisms inbuilt, and are naturally able to adapt to these new incoming data streams. It can also support active learning paradigm \cite{laptev2015generic}, which allows models to interactively query users for labels on data points for which it is uncertain about, and subsequently update the model with the new labels.



\paragraph{Dimensionality} 
Each metric of monitoring data forms a univariate time series, and thus a service usually contains multiple metrics, each of which describes a different part or attribute of a complex entity, constituting a multivariate time series. The conventional solution is to build univariate time series anomaly detection for each metric. However, for a complex system, it ignores the intrinsic interactions among each metric and cannot well represent the system's overall status. Naively combining the anomaly detection results of each univariate time series performs poorly for multivariate anomaly detection method \cite{DBLP:conf/kdd/LiZHSJWP21}, since it cannot model the inter-dependencies among metrics for a service.

\textbf{Model}
A wide range of machine learning models can be used for time series anomaly detection, broadly classified as deep learning models, tree-based models, and statistical models.
Deep learning models \cite{audibert2020usad, xu2018unsupervised, su2019robust, li2021multivariate, huang2022semi, yang2022causality, ayed2020anomaly, rabanser2022intrinsic} leverage the success and power deep neural networks to learn representations of the time series data. These representations of time series data contain rich semantic information of the underlying metric, and can be used as a  reconstruction-based, unsupervised method. Tree-based methods leverage a tree structure as a density-based, unsupervised method \cite{guha2016robust}.
Statistical models \cite{hochenbaum2017automatic} rely on classical statistical tests, which are considered a reconstruction-based method.


\textbf{Industrial Practices} 
Building a system which can handle the large amounts of metric data generated in real cloud IT operations is often an issue. This is because the metric data in real-world scenarios is quite diverse and the definition of anomaly may vary in different scenarios. Moreover, almost all time series anomaly detection systems require to handle a large amount of metrics in parallel with low-latency \cite{gao2020robusttad}. Thus, works which propose a system to handle the infrastructure are highlighted here. 
EGADS \cite{laptev2015generic} is a system by Yahoo!, scaling up to millions of data points per second, and focuses on optimizing real-time processing. It comprises a batch time series modelling module, an online anomaly detection module, and an alerting module. It leverages a variety of unsupervised methods for anomaly detection, and an optional active learning component for filtering alerts.
\cite{ren2019time} is a system by Microsoft, which includes three major components, a data ingestion, experimentation, and online compute platform. They propose an efficient deep learning anomaly detector to achieve high accuracy and high efficiency at the same time.
\cite{gao2020robusttad} is a system by Alibaba group, comprising data ingestion, offline training, online service, and visualization and alarms modules. They propose a robust anomaly detector by using time series decomposition, and thus can easily handle time series with different characteristics, such as different seasonal length, different types of trends, etc.
\cite{huang2022semi} is a system by Tencent, comprising of a offline model training component and online serving component, which employs active learning to update the online model via a small number of uncertain samples.

\vspace{1em}
\textbf{Challenges}

\textbf{Lack of labels}
The main challenge of metric anomaly detection is the lack of ground truth anomaly labels \cite{DBLP:journals/csur/SoldaniB23, DBLP:conf/kdd/LiZHSJWP21}. Due to the open-ended nature and complexity of incidents in server architectures, it is difficult to define what an anomaly is. Thus, building labelled datasets is an extremely labor and resource intensive exercise, one which requires the effort of domain experts to identify anomalies from time series data. Furthermore, manual labelling could lead to labelling errors as there is no unified and formal definition of an anomaly, leading to subjective judgements on ground truth labels \cite{DBLP:journals/corr/abs-2009-13807}.

\textbf{Real-time inference}
A typical cloud infrastructure could collect millions of data points in a second, requiring near real-time inference to detect anomalies. Metric anomaly detection systems need to be scalable and efficient \cite{DBLP:conf/ipccc/BuLZMLZP18, DBLP:journals/csur/SoldaniB23}, optionally supporting model retraining, leading to immense compute, memory, and I/O loads. The increasing complexity of anomaly detection models with the rising popularity of deep learning methods \cite{DBLP:journals/corr/abs-2211-05244} add a further strain on these systems due to the additional computational cost these larger models bring about. 

\textbf{Non-stationarity of metric streams}
The temporal patterns of metric data streams typically change over time as they are generated from non-stationary environments \cite{DBLP:conf/icml/Huang0GG19}. The evolution of these patterns is often caused by exogenous factors which are not observable. One such example is that the growth in the popularity of a service would cause customer metrics (e.g. request count) to drift upwards over time.
Ignoring these factors would cause a deterioration in the anomaly detector's performance. One solution is to continuously update the model with the recent data \cite{DBLP:journals/corr/abs-2202-11672}, but this strategy requires carefully balancing of the cost and model robustness with respect to the updating frequency.

\textbf{Public benchmarks}
While there exists benchmarks for general anomaly detection methods and time series anomaly detection methods \cite{han2022adbench, DBLP:conf/nips/LaiZXZWH21}, there is still a lack of benchmarking for metric incident detection in AIOps domain.
Given the wide and diverse nature of time series data, they often exhibit a mixture of different types of anomaly depends on specific domain, making it challenging to understand the pros and cons of algorithms \cite{DBLP:conf/nips/LaiZXZWH21}. Furthermore, existing datasets have been criticised to be trivial and mislabelled \cite{wu2021current}.

\vspace{1em}
\textbf{Future Trends}

\textbf{Active learning/human-in-the-loop}
To address the problem of lacking of labels, a more intelligent way is to integrate human knowledge and experience with minimum cost. As special agents, humans have rich prior knowledge \cite{DBLP:journals/fgcs/WuXSZM022}. If the incident detection framework can encourage the machine learning model to  engage with learning operation expert wisdom and knowledge, it would help deal with scarce and noise label issue. The use of active learning to update online model in \cite{huang2022semi} is a typical example to incorporate human effort in the annotation task. There are certainly large research scope for incorporating human effort in other data processing step, like feature extraction. Moreover, the human effort can also be integrated in the machine learning model training and inference phase.


\textbf{Streaming updates}
Due to the non-stationarity of metric streams, keeping the anomaly detector updated is of utmost importance. Alongside the increasingly complex models and need for cost-effectiveness, we will see a move towards methods with the built-in capability of efficient streaming updates. With the great success of deep learning methods in time series anomaly detection tasks \cite{choi2021deep}. Online deep learning is an increasingly popular topic \cite{DBLP:journals/corr/abs-1711-03705}, and we may start to see a transference of techniques into metric anomaly detection for time-series in the near future.

\textbf{Intrinsic anomaly detection}
Current research works on time series anomaly detection do not distinguish the cause or the type of anomaly, which is critical for the subsequent mitigation steps in AIOps. For example, even anomaly are successfully detected, which is caused by extrinsic environment, the operator is unable to mitigate its negative effect. Introduced in \cite{rabanser2022intrinsic, yang2022causality}, intrinsic anomaly detection considers the functional dependency structure between the monitored metric, and the environment. This setting considers changes in the environment, possibly leveraging information that may not be available in the regular (extrinsic) setting. For example, when scaling up/down the resources serving an application (perhaps due to autoscaling rules), we will observe a drop/increase in CPU metric. While this may be considered as an anomaly in the extrinsic setting, it is in fact not an incident and accordingly, is not an anomaly in the intrinsic setting.

%% file: log_ad.tex
\vspace{1em}
\textbf{Problem Definition} 

Software and system logging data is one of the most popular ways of recording and tracking runtime information about all ongoing processes within a system, to any arbitrary level of granularity. Overall, a large distributed system can have massive volume of heterogenous logs dumped by its different services or microservices, each having time-stamped text messages following their own unstructured or semi-structured or structured format. Throughout various kinds of IT Operations these logs have been widely used by reliability and performance engineers as well as core developers in order to understand the system’s internal
status and to facilitate monitoring, administering, and troubleshooting \cite{10.1007/s11036-021-01832-3,10.1145/3460345,10.1145/3483424,10.1145/3459637.3482209,10.1145/3510003.3510155,10.1145/3468264.3473933,DBLP:journals/corr/abs-2112-03159,10.1145/3501297,https://doi.org/10.48550/arxiv.2107.05908}. More, specifically, in the AIOps pipeline, one of the foremost tasks that log analysis can cater to is log based Incident Detection. This is typically achieved through anomaly detection over logs which aims to detect the anomalous loglines or sequences of loglines that indicate possible occurrence of an incident, from the humungous amounts of software logging data dumps generated by the system. Log based anomaly detection is generally applied once an incident has been detected based on monitoring of KPI metrics, as a more fine-grained incident detection or failure diagnosis step in order to detect which service or micro-service or which software module of the system execution is behaving anomalously. 

\vspace{1em}
\textbf{Task Complexity} 

\emph{Diversity of Log Anomaly Patterns}: There are very diverse kinds of incidents in AIOps which can result in different kinds of anomaly patterns in the log data - either manifesting in the log template (i.e. the constant part of the log line) or the log parameters (i.e. the variable part of the log line containing dynamic information). These are 
{i) keywords} - appearance of keywords in log lines bearing domain-specific semantics of failure or incident or abnormality in the system (e.g. out of memory or crash)
{ii) template count} - where a sudden increase or decrease of log templates or log event types is indicative of anomaly
{iii) template sequence} - where some significant deviation from the normal order of task execution is indicative of anomaly
{iv) variable value} - some variables associated with some log templates or events can have physical meaning (e.g. time cost) which could be extracted out and aggregated into a structured time series on which standard anomaly detection techniques can be applied. 
{v) variable distribution} - for some categorical or numerical variables, a deviation from the standard distribution of the variable can be indicative of an anomaly
{vi) time interval} - some performance issues may not be explicitly observed in the logline themselves but in the time interval between specific log events. 
\vspace{0.5em}

\emph{Need for AI}: Given the humongous nature of the logs, it is often infeasible for even domain experts to manually go through the logs to detect the anomalous loglines. Additionally, as described above, depending on the nature of the incident there can be diverse types of anomaly patterns in the logs, which can manifest as anomalous key words (like "errors" or "exception") in the log templates or the volume of specific event logs or distribution over log variables or the time interval between two log specific event logs. However, even for a domain expert it is not possible to come up with rules to detect these anomalous patterns, and even when they can, they would likely not be robust to diverse incident types and changing nature of log lines as the software functionalities change over time. Hence, this makes a compelling case for employing data-driven models and machine intelligence to mine and analyze this complex data-source to serve the end goals of incident detection. 

\begin{figure*}[htbp!]
    \centering
    \includegraphics[width=\linewidth]{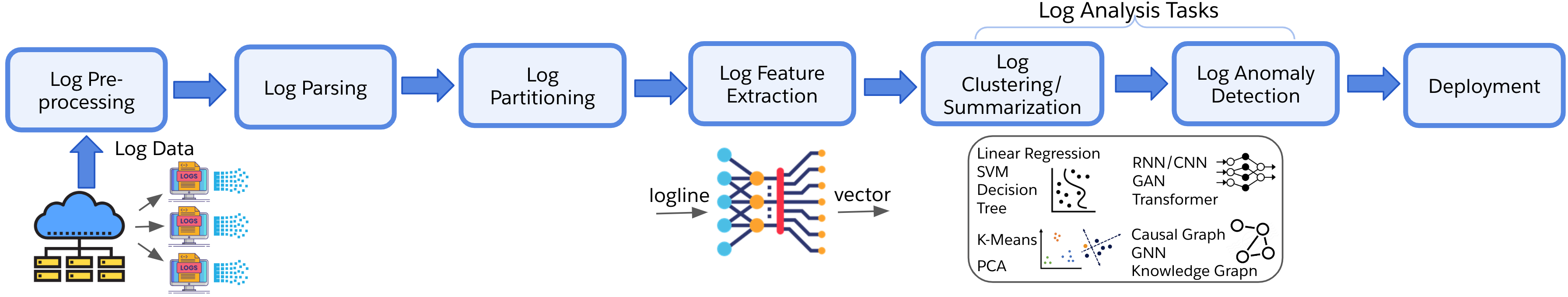}
    \caption{Steps of the Log Analysis Workflow for Incident Detection}
    \label{fig:log_ad_workflow}
\end{figure*}

\vspace{1em}
\textbf{Log Analysis Workflow for Incident Detection}

In order to handle the complex nature of the data, typically a series of steps need to be followed to meaningfully analyze logs to detect incidents. Starting with the raw log data or data streams, the log analysis workflow first does some preprocessing of the logs to make them amenable to ML models. This is typically followed by log parsing which extracts a loose structure from the semi-structured data and then grouping and partitioning the log lines into log sequences in order to model the sequence characteristics of the data. After this, the logs or log sequences are represented as a machine-readable matrix on which various log analysis tasks can be performed - like clustering and summarizing the huge log dumps into a few key log patterns for easy visualization or for detecting anomalous log patterns that can be indicative of an incident. Figure \ref{fig:log_ad_workflow} provides an outline of the different steps in the log analysis wokflow. While some of these steps are more of engineering challenges, others are more AI-driven and some even employ a combination of machine learning and domain knowledge rules.

\vspace{1em}
\textit{i) Log Preprocessing:} This step typically involves customised filtering of specific regular expression patterns (like IP addresses or memory locations) that are deemed irrelevant for the actual log analysis. Other preprocessing steps like tokenization requires specialized handling of different wording styles and patterns arising due to the hybrid nature of logs consisting of both natural language and programming language constructs. For example a log line can contain a mix of text strings from source-code data having snake-case and camelCase tokens along with white-spaced tokens in natural language.  
 
\begin{figure}[htbp!]
    \centering
    \fbox{\includegraphics[width=0.8\linewidth]{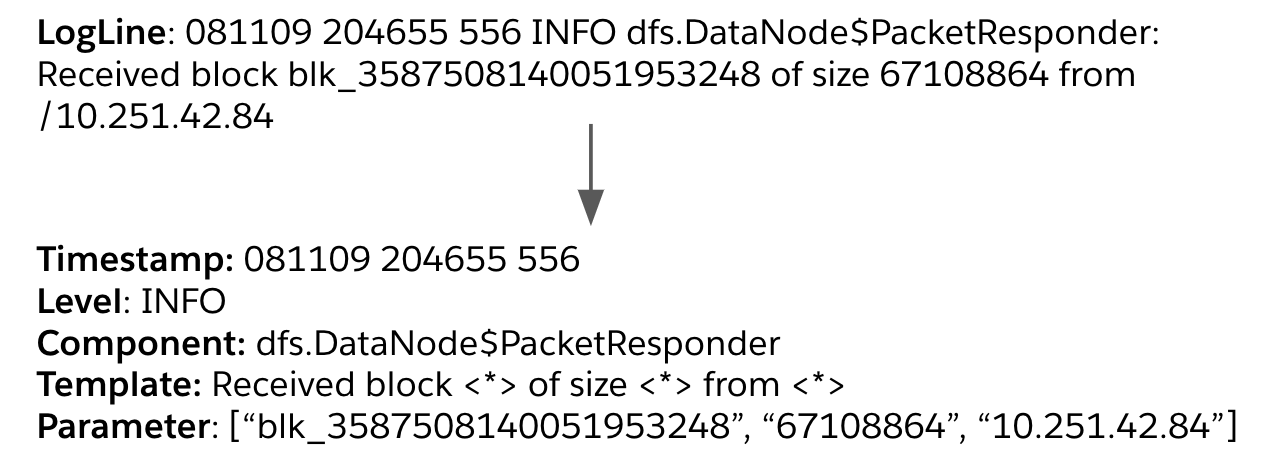}}
    \caption{Example of Log Parsing}
    \label{fig:log_parsing}
\end{figure}

\vspace{1em}
\textit{ii) Log Parsing:} To enable downstream processing, unstructured log messages first need to be parsed into a structured event template (i.e. constant part that was actually designed by the developers) and parameters (i.e. variable part which contain the dynamic runtime information). Figure \ref{fig:log_parsing} provides one such example of parsing a single log line. In literature there have been heuristic methods for parsing as well as AI-driven methods which include traditional ML and also more recent neural models. The heuristic methods like Drain \cite{8029742}, IPLoM \cite{10.1145/1557019.1557154} and AEL \cite{4601543} exploit known inductive bias on log structure while Spell \cite{7837916} uses Longest common subsequence algorithm to dynamically extract log patterms. Out of these, Drain and Spell are most popular, as they scale well to industrial standards. Amongst the traditional ML methods, there are i) Clustering based methods like LogCluster \cite{DBLP:conf/cnsm/VaarandiP15}, LKE \cite{5360240}, LogSig \cite{10.1145/2063576.2063690}, SHISO \cite{6649746}, LenMa \cite{DBLP:journals/corr/Shima16}, LogMine \cite{10.1145/2983323.2983358} which assume that log message types coincide in similar groups ii) Frequent pattern mining and item-set mining methods SLCT \cite{1251233}, LFA \cite{5463281} to extract common message types iii) Evolutionary optimization approaches like MoLFI \cite{10.1145/3196321.3196340}. On the other hand, recent neural methods include \cite{10.1007/978-3-030-67667-4_8} - Neural Transformer based models which use self-supervised Masked Language Modeling to learn log parsing vii) UniParser \cite{10.1145/3485447.3511993} - an unified parser for heterogenous log data with a learnable similarity module to generalize to diverse logs across different systems.
There are yet another class of log analysis methods \cite{9678773,DBLP:journals/corr/abs-2111-09564} which aim at parsing free techniques, in order to avoid the computational overhead of parsing and the errors cascading from erroneous parses, especially due to the lack of robustness of the parsing methods. 

\begin{figure}[htbp!]
    \centering
    \fbox{\includegraphics[width=0.7\linewidth]{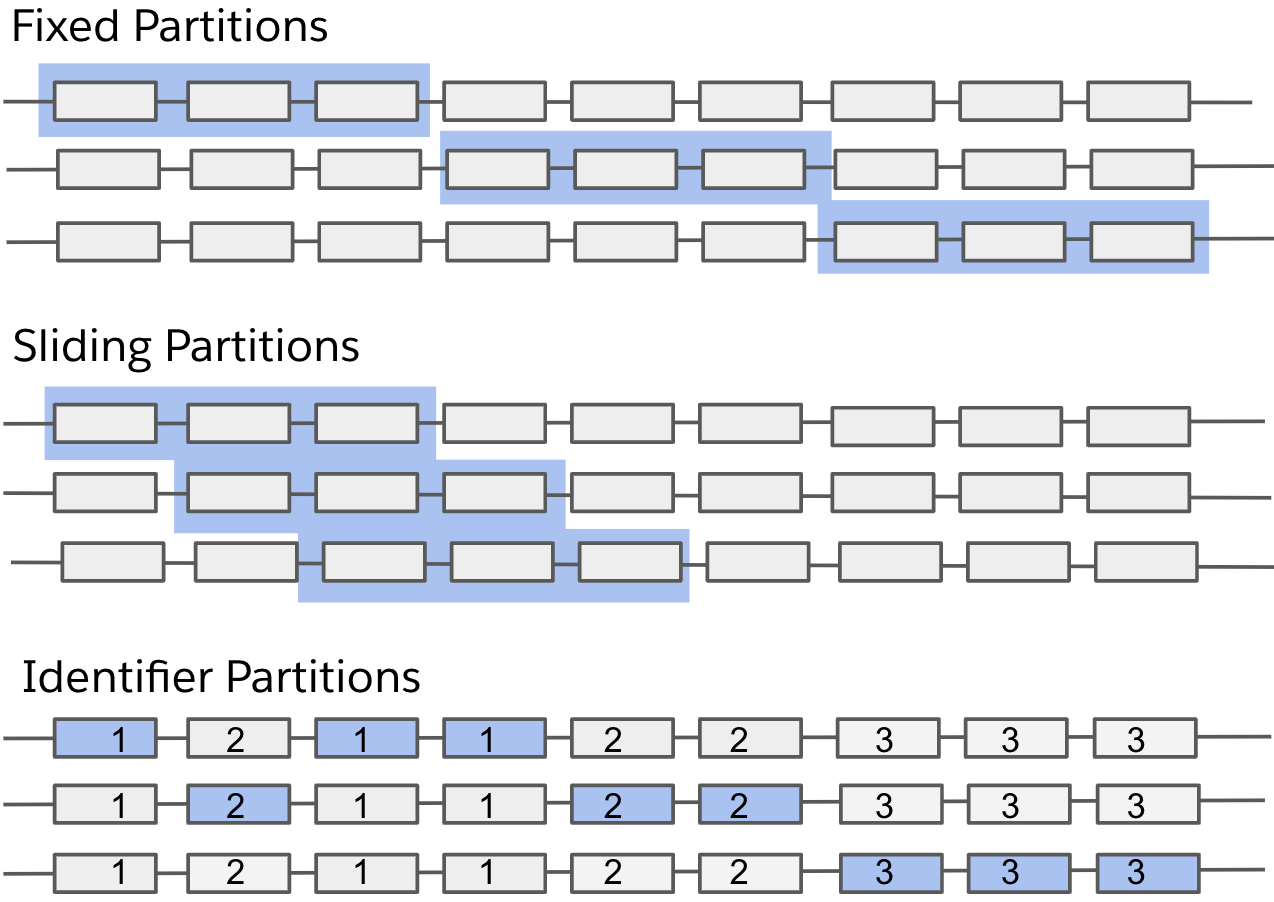}}
    \caption{Different types of log partitioning}
    \label{fig:log_partition}
\end{figure}

\vspace{1em}
\textit{iii) Log Partitioning:} After parsing the next step is to partition the log data into groups, based on some semantics where each group represents a finite chunk of log lines or log sequences. The main purpose behind this is to decompose the original log dump typically consisting of millions of log lines into logical chunks, so as to enable explicit modeling on these chunks and allow the models to capture anomaly patterns over sequences of log templates or log parameter values or both. Log partitioning can be of different kinds \cite{10.1145/3468264.3473933,7381796} - Fixed or Sliding window based partitions, where the length of window is determined by length of log sequence or a period of time, and Identifier based partitions where logs are partitioned based on some identifier (e.g. the session or process they originate from). Figure \ref{fig:log_partition} illustrates these different choices of log grouping and partitioning. A log event is eventually deemed to be anomalous or not, either at the level of a log line or a log partition. 

\vspace{1em}
\textit{iv) Log Representation:} After log partitioning, the next step is to represent each partition in a machine-readable way (e.g. a vector or a matrix) by extracting features from them. This can be done in various ways \cite{10.5555/3367471.3367702,7381796}- either by extracting specific handcrafted features using domain knowledge or through ii) sequential representation which converts each partition to an ordered sequence of log event ids ii) quantitative representation which uses count vectors, weighted by the term and inverse document frequency information of the log events iii) semantic representation captures the linguistic meaning from the sequence of language tokens in the log events and learns a high-dimensional embedding vector for each token in the dataset. The nature of log representation chosen has direct consequence in terms of which patterns of anomalies they can support - for example, for capturing keyword based anomalies, semantic representation might be key, while for anomalies related to template count and variable distribution, quantitative representations are possibly more appropriate. The semantic embedding vectors themselves can be either obtained using pretrained neural language models like GloVe, FastText, pretrained Transformer like BERT, RoBERTa etc or learnt using a trainable embedding layer as part of the target task. 

\vspace{1em}
\textit{v) Log Analysis tasks for Incident Detection:} Once the logs are represented in some compact machine-interpretable form which can be easily ingested by AI models, a pipeline of log analysis tasks can be performed on it - starting with Log compression techniques using Clustering and Summarization, followed by Log based Anomaly Detection. In turn, anomaly detection can further enable downstream tasks in Incident Management like Failure Prediction and Root Cause Analysis. In this section we discuss only the first two log analysis tasks which are pertinent to incident detection and leave failure prediction and RCA for the subsequent sections.

\vspace{1em}
\indent \textit{v.1) Log Compression through Clustering \& Summarization:} 
This is a practical first-step towards analyzing the huge volumes of log data is Log Compression through various clustering and summarization techniques. The objective of this analysis serves two purposes - Firstly, this step can independently help the site reliability engineers and service owners during incident management by providing a practical and intuitive way of visualizing these massive volumes of complex unstructured raw log data. Secondly, the output of log clustering can directly be leveraged in some of the log based anomaly detection methods. 

Amongst the various techniques of log clustering, \cite{10.1145/2889160.2889232,DBLP:conf/cnsm/VaarandiP15,DBLP:journals/ejwcn/YangQQDZ19} employ hierarchical clustering and can support online settings by constructing and retrieving from knowledge base of representative log clusters. \cite{DBLP:conf/sosp/XuHFPJ09,10.1007/978-3-642-54903-8_38} use frequent pattern matching with dimension reduction techniques like PCA and locally sensitive hashing with online and streaming support. \cite{10.1109/ASE.2019.00085,10.1145/1557019.1557154,10.1145/3098954.3098973} uses efficient iterative or incremental clustering and partitioning techniques that support online and streaming logs and can also handle clustering of rare log instances. Another area of existing literature \cite{4354137,DBLP:journals/corr/abs-2012-08938,DIJKMAN2017114,9442364} focus on log compression through summarization - where, for example, \cite{4354137} uses heuristics like log event ids and timings to summarize and   \cite{DBLP:journals/corr/abs-2012-08938,DBLP:journals/corr/abs-2112-03159} does openIE based triple extraction using semantic information and domain knowledge and rules to generate summaries, while \cite{DIJKMAN2017114,9442364} use sequence clustering using linguistic rules or through grouping common event sequences.

\vspace{1em}
\indent \textit{v.2) Log Anomaly Detection:} Perhaps the most common use of log analysis is for log based anomaly detection where a wide variety of models have been employed in both research and industrial settings. These models are categorized based on various factors i) the learning setting - supervised, semi-supervised or unsupervised: While the semi-supervised models assume partial knowledge of labels or access to few anomalous instances, unsupervised ones train on normal log data and detect anomaly based on their prediction confidence. ii) the type of Model - Neural or traditional statistical non-neural models iii) the kinds of log representations used iv) Whether to use log parsing or parser free methods v) If using parsing, then whether to encode only the log template part or both template and parameter representations iv) Whether to restrict modeling of anomalies at the level of individual log lines or to support sequential modeling of anomaly detection over log sequences. 

The nature of log representation employed and the kind of modeling used - both of these factors influence what type of anomaly patterns can be detected - for example keyword and variable value based anomalies are captured by semantic representation of log lines, while template count and variable distribution based anomaly patterns are more explicitly modeled through quantitative representations of log events. Similarly template sequence and time-interval based anomalies need sequential modeling algorithms which can handle log sequences. 

Below we briefly summarize the body of literature dedicated to these two types of models - Statistical and Neural; and In Table \ref{tab:log_ad_comp} we provide a comparison of a more  comprehensive list of existing anomaly detection algorithms and systems. 

Statistical Models are the more traditional machine learning models which draw inference from various statistics underlying the training data. In the literature there have been various statistical ML models employed for this task under different training settings. Amongst the supervised methods,  \cite{10.1145/1755913.1755926,4470294,1301345} using traditional learning strategies of Linear Regression, SVM, Decision Trees, Isolation Forest with handcrafted features extracted from the entire logline. Most of these model the data at the level of individual log-lines and cannot not explicitly capture sequence level anomalies.  There are also unsupervised methods like ii) dimension reduction techniques like Principal Component Analysis (PCA) \cite{DBLP:conf/sosp/XuHFPJ09} iii) clustering and drawing correlations between log events and metric data as in \cite{DBLP:conf/cnsm/VaarandiP15,10.1145/2889160.2889232,10.1145/3236024.3236083,7381796}. There are also unsupervised pattern mining methods which include  mining invariant patterns from singular value decomposition \cite{DBLP:conf/usenix/LouFYXL10} and mining frequent patterns from execution flow and control flow graphs \cite{10.1145/2939672.2939712,8029741,8030620,5360240}. Apart from these there are also systems which employ a rule engine built using domain knowledge and an ensemble of different ML models to cater to different incident types \cite{10.1145/3468264.3473933} and also heuristic methods for doing contrast analysis between normal and incident-indicating abnormal logs \cite{10.1145/3468264.3473919}. 

Neural Models, on the other hand are a more recent class of machine learning models which use artificial neural networks and have proven remarkably successful across numerous AI applications. They are particularly powerful in encoding and representing the complex semantics underlying in a way that is meaningful for the predictive task. One class of unsupervised neural models use reconstruction based self-supervised techniques to learn the token or line level representation, which includes i)  Autoencoder models \cite{9209707,FARZAD2020229} ii) more powerful self-attention based Transformer models \cite{9338283} iv) specific pretrained Transformers like BERT language model \cite{9401970,9534113,DBLP:journals/corr/abs-2112-03159}. Another offshoot of reconstruction based models is those using generative adversarial or GAN paradigm of training for e.g. \cite{10.1007/s10796-020-10026-3,10.1007/s10489-021-02863-9} using LSTM or Transformer based encoding. The other types of unsupervised models are forecasting based, which learn to predict the next log token or next log line in a self-supervised way - for e.g i) Recurrent Neural Network based models like LSTM \cite{10.1145/3133956.3134015,9527018,9251092,10.1145/3459637.3482209,10.1145/3338906.3338931} and GRU \cite{9401970} or their attention based counterparts \cite{10.5555/3367471.3367702,10.1145/3217871.3217872,9251078} ii) Convolutional Neural Network (CNN)  based models \cite{DBLP:conf/dasc/LuWLW18} or more complex models which use Graph Neural Network to represent log event data \cite{https://doi.org/10.48550/arxiv.2209.07869,10.1007/978-3-030-75762-5_6}. Both reconstruction and forecasting based models are capable of handling sequence level anomalies, it depends on the nature of training (i.e. whether representations are learnt at log line or token level) and the capacity of model to handle long sequences (e.g. amongst the above, Autoencoder models are the most basic ones). 

Most of these models follow the practical setup of unsupervised training, where they train only non-anomalous log data. However, other works have also focused on supervised training of LSTM, CNN and Transformer models \cite{10.1145/3338906.3338931,DBLP:conf/dasc/LuWLW18,9678773,9244088}, over anomalous and normal labeled data. On the other hand, \cite{9401970,9251092} use weak supervision based on heuristic assumptions for e.g. logs from external systems are considered anomalous. Most of the neural models use semantic token representations, some with pretrained fixed or trainable embeddings, initialized with GloVe, fastText or pretrained transformer based models, BERT, GPT, XLM etc. 

\vspace{1em}

\textit{vi) Log Model Deployment:} The final step in the log analysis workflow is deployment of these models in the actual industrial settings. It involves i) a training step, typically over offline log data dump, with or without some supervision labels collected from domain experts ii) online inference step, which often needs to handle practical challenges like non-stationary streaming data i.e. where the data distribution is not independently and identically distributed throughout the time. For tackling this, some of the more traditional statistical methods like \cite{9338283,10.1145/3236024.3236083,10.1145/2889160.2889232,DBLP:conf/sosp/XuHFPJ09} support online streaming update while some other works can also adapt to evolving log data by incrementally building a knowledge base or memory or out-of-domain vocabulary \cite{9209707}. On the other hand most of the unsupervised models support syncopated batched online training, allowing the model to continually adapt to changing data distributions and to be deployed on high throughput streaming data sources. However for some of the more advanced neural models, the online updation might be too computationally expensive even for regular batched updates. 

Apart from these, there have also been specific work on other challenges related to model deployment in practical settings like transfer learning across logs from different domains or applications \cite{9251092,9338283,10.1145/3459637.3482209,10.1145/3459637.3482209,https://doi.org/10.48550/arxiv.2210.04145} under semi-supervised settings using only supervision from source systems. Other works focus on evaluating model robustness and generalization (i.e. how well the model adapts to) to unstable log data due to continuous logging modifications throughout software evolutions and updates \cite{9527018,10.1145/3338906.3338931,9401970}. They achieve these by adopting domain adversarial paradigms during training \cite{10.1145/3459637.3482209,10.1145/3459637.3482209} or using counterfactual explanations \cite{https://doi.org/10.48550/arxiv.2210.04145} or multi-task settings \cite{DBLP:journals/corr/abs-2112-03159} over various log analysis tasks.





\vspace{1em}
\textbf{Challenges \& Future Trends}
\vspace{0.5em}
\paragraph*{Collecting supervision labels} Like most AIOps tasks, collecting large-scale supervision labels for training or even evaluation of log analysis problems is very challenging and impractical as it involves significant amount of manual intervention and domain knowledge. For log anomaly detection, the goal being quite objective, label collection is still possible to enable atleast a reliable evaluation. Whereas, for other log analysis tasks like clustering and summarization, collecting supervision labels from domain experts is often not even possible as the goal is quite subjective and hence these tasks are typically evaluated through the downstream log analysis or RCA task. 
\vspace{0.5em}
\paragraph*{Imbalanced class problem} One of the key challenges of anomaly detection tasks, is the class imbalance, stemming from the fact that anomalous data is inherently extremely rare in occurrence. Additionally, various systems may show different kinds of data skewness owing to the diverse kinds of anomalies listed above.  This poses a technical challenge both during model training with highly skewed data as well as choice of evaluation metrics, as Precision, Recall and F-Score may not perform satisfactorily. Further at inference, thresholding over the anomaly score gets particularly challenging for unsupervised models. While for benchmarking purposes, evaluation metrics like AUROC (Area under ROC curve) can suffice, but for practical deployment of these models require either careful calibrations of anomaly scores or manual tuning or heuristic means for setting the threshold. This being quite sensitive to the application at hand, also poses realistic challenges when generalizing to heterogenous logs from different systems.
\vspace{0.5em}
\paragraph*{Handling large volume of data} Another challenge in log analysis tasks is handling the huge volumes of logs, where most large-scale cloud-based systems can generate petabytes of logs each day or week. This calls for log processing algorithms, that are not only effective but also lightweight enough to be very fast and efficient.
\vspace{0.5em}
\paragraph*{Handling non-stationary log data} Along with humongous volume, the natural and most practical setting of logs analysis is an online streaming setting, involving non-stationary data distribution - with heterogenous log streams coming from different inter-connected micro-services, and the software logging data itself evolving over time as developers naturally keep evolving software in the agile cloud development environment. This requires efficient online update schemes for the learning algorithms and specialized effort towards building robust models and evaluating their robustness towards unstable or evolving log data.
\vspace{0.5em}
\paragraph*{Handling noisy data} Annotating log data being extremely challenging even for domain experts, supervised and semi-supervised models need to handle this noise during training, while for unsupervised models, it can heavily mislead evaluation. Even though it affects a small fraction of logs, the extreme class imbalance aggrevates this problem. Another related challenge is that of errors compounding and cascading from each of the processing steps in the log analysis workflow when performing the downstream tasks like anomaly detection.
\vspace{0.5em}
\paragraph*{Realistic public benchmark datasets for anomaly detection} Amongst the publicly available log anomaly detection datasets, only a limited few contain anomaly labels. Most of those benchmarks have been excessively used in the literature and hence do not have much scope of furthering research. Infact, their biggest limitation is that they fail to showcase the diverse nature of incidents that typically arise in real-world deployment. Often very simple handcrafted rules prove to be quite successful in solving anomaly detection tasks on these datasets. Also, the original scale of these datasets are several orders of magnitude smaller than the real-world use-cases and hence not fit for showcasing the challenges of online or streaming settings. Further, the volume of unique patterns collapses significantly after the typical log processing steps to remove irrelevant patterns from the data. On the other hand, a vast majority of the literature is backed up by empirical analysis and evaluation on internal proprietary data, which cannot guarantee reproducibility. This calls for more realistic public benchmark datasets that can expose the real-world challenges of aiops-in-the-wild and also do a fair benchmarking across contemporary log analysis models.
\vspace{0.5em}
\paragraph*{Public benchmarks for parsing, clustering, summarization} Most of the log parsing, clustering and summarization literature only uses a very small subset of data from some of the public log datasets, where the oracle parsing is available, or in-house log datasets from industrial applications where they compare with oracle parsing methods that are unscalable in practice. This also makes fair comparison and standardized benchmarking difficult for these tasks.
\vspace{0.5em}
\paragraph*{Better log language models} Some of the recent advances in neural NLP models like transformer based language models BERT, GPT has proved quite promising for representing logs in natural language style and enabling various log analysis tasks. However there is more scope of improvement in building neural language models that can appropriately encode the semi-structured logs composed of fixed template and variable parameters without depending on an external parser.
\vspace{0.5em}
\paragraph*{Incorporating Domain Knowledge} While existing log anomaly detection systems are entirely rule-based or automated, given the complex nature of incidents and the diverse varieties of anomalies, a more practical approach would involve incorporating domain knowledge into these models either in a static form or dynamically, following a human-in-the-loop feedback mechanism. For example, in a complex system generating humungous amounts of logs, which kinds of incidents are more severe and which types of logs are more crucial to monitor for which kind of incidents. Or even at the level of loglines, domain knowledge can help understand the real-world semantics or physical significance of some of the parameters or variables mentioned in the logs. These aspects are often hard for the ML system to gauge on its own especially in the practical unsupervised settings.  
\vspace{0.5em}
\paragraph*{Unified models for heterogenous logs} Most of the log analysis models are highly sensitive towards the nature of log preprocessing or grouping, needing customized preprocessing for each type of application logs. This alludes towards the need for unified models with more generalizable preprocessing layers that can handle heterogenous kinds of log data and also different types of log analysis tasks. While \cite{DBLP:journals/corr/abs-2112-03159} was one of the first works to explore this direction, there is certainly more research scope for building practically applicable models for log analysis.

%% file: trace_ad.tex
\vspace{1em}
\textbf{Problem Definition}

Traces are semi-structured event logs with span information about the topological structure of the service graph. Trace anomaly detection relies on finding abnormal paths on the topological graph at given moments, as well as discovering abnormal information directly from trace event log text. There are multiple ways to process trace data. Traces usually have timestamps and associated sequential information so it can be covered into time-series data. Traces are also stored as trace event logs, containing rich text information. Moreover, traces store topological information which can be used to reconstruct the service graphs that represents the relation among components of the systems. From the data perspective, traces can easily been turned into multiple data modalities. Thus, we combines trace-based anomaly detection with multi-modal anomaly detection to discuss in this section. Recently, we can see with the help of multi-modal deep learning technologies, trace anomaly detection can combine different levels of information relayed by trace data and learn more comprehensive anomaly detection models \cite{Nedelkoski2019}\cite{Zhang2022}. 

\vspace{1em}
\textbf{Empirical Approaches}

Traces draw more attention in microservice system architectures since the topological structure becomes very complex and dynamic. Trace anomaly detection started from practical usages for large scale system debugging \cite{Arnold2007}. Empirical trace anomaly detection and RCA started with constructing trace graphs and identifying abnormal structures on the constructed graph. Constructing the trace graph from trace data is usually very time consuming, an offline component is designed to train and construct such trace graph. Apart from , to adapt to the usage requirements to detect and locate issues in large scale systems, trace anomaly detection and RCA algorithms usually also have an online part to support real-time service. For example, Cai \textit{et al.}. released their study of a real-time trace-level diagnosis system, which is adopted by Alibaba datacenters. This is one of the very few studies to deal with real large distributed systems \cite{cai2019real}.

Most empirical trace anomaly detection work follow the offline and online design pattern to construct their graph models. In the offline modeling, unsupervised or semi-supervised techniques are utilized to construct the trace entity graphs, very similar to techniques in process discovery and mining domain. For example, PageRank has been used to construct web graphs in one of the early web graph anomaly detection works \cite{Papadimitriou2010}.  After constructing the trace entity graphs, a variety of techniques can be used to detect anomalies. One common way is to compare the current graph pattern to normal graph patterns. If the current graph pattern significantly deviates from the normal patterns, report anomalous traces. 

An alternative approach is using data mining and statistical learning techniques to run dynamic analysis without constructing the offline trace graph. Chen \textit{et al.} proposed Pinpoint \cite{Chen2002}, a framework for root cause analysis that using coarse-grained tagging data of real client requests at real-time when these requests traverse through the system, with data mining techniques. Pinpoint discovers the correlation between success / failure status of these requests and fault components. The entire approach processes the traces on-the-fly and does not leverage any static dependency graph models. 

\vspace{1em}
\textbf{Deep Learning Based Approaches}

In recent years, deep learning techniques started to be employed in trace anomaly detection and RCA. Also with the help of deep learning frameworks, combining general trace graph information and the detailed information inside of each trace event to train multimodal learning models become possible. 

Long-short term memory (LSTM) network \cite{Hochreiter1997} is a very popular neural network model in early trace and multimodal anomaly detection. LSTM is a special type of recurrent neural network (RNN) and has been proved to success in lots of other domains. In AIOps, LSTM is also commonly used in metric and log anomaly detection applications. Trace data is a natural fit with RNNs, majorly in two ways: 1) The topological order of traces can be modeled as event sequences. These event sequences can easily be transformed into model inputs of RNNs. 2) Trace events usually have text data that conveys rich information. The raw text, including both the structured and unstructured parts, can be transformed into vectors via standard tokenization and embedding techniques, and feed the RNN as model inputs. Such deep learning model architectures can be extended to support multimodal input, such as combining trace event vector with numerical time series values \cite{Nedelkoski2019}. 

To better leverage the topological information of traces, graph neural networks have also been introduced in trace anomaly detection. Zhang \textit{et al.} developed DeepTraLog, a trace anomaly detection technique that employs Gated graph neural networks \cite{Zhang2022}. DeepTraLog targets to solve anomaly detection problems for complex microservice systems where service entity relationships are not easy to obtain. Moreover, the constructed graph by GGNN training can also be used to localize the issue, providing additional root-cause analysis capability. 

\vspace{1em}
\textbf{Limitations}

Trace data became increasingly attractive as more applications transitioned from monolithic to microservice architecture. There are several challenges in machine learning based trace anomaly detection.

\textbf{Data quality.} As far as we know, there are multiple trace collection platforms and the trace data format and quality are inconsistent across these platforms, especially in the production environment. To use these trace data for analysis, researchers and developers have to spend significant time and effort to clean and reform the data to feed machine learning models. 

\textbf{Difficult to acquire labels.} It is very difficult to acquire labels for production data. For a given incident, labeling the corresponding trace requires identifying the incident occurring time and location, as well as the root cause which may be located in totally different time and location. Obtaining such full labels for thousands of incidents is extremely difficult. Thus, most of the existing trace analysis research still use synthetic data to evaluate the model performance. This brings more doubts whether the proposed solution can solve problems in real production.

\textbf{No sufficient multimodal and graph learning models.} Trace data are complex. Current trace analysis simplifies trace data into event sequences or time-series numerical values, even in the multimodal settings. However, these existing model architectures did not fully leverage all information of trace data in one place. Graph-based learning can potentially be a solution but discussions of this topic are still very limited.

\textbf{Offline model training.} The deep learning models in existing research relies on offline model training, partially because model training is usually very time consuming and contradicts with the goal of real-time serving. However, offline model training brings static dependencies to a dynamic system. Such dependencies may cause additional performance issues.

\vspace{1em}
\textbf{Future Trends}

\textbf{Unified trace data} Recently, OpenTelemetry leads the effort to unify observability telemetry data, including metrics, logs, traces, etc., across different platforms. This effort can bring huge benefits to future trace analysis. With more unified data models, AI researchers can more easily acquire necessary data to train better models. The trained model can also be easily plug-and-play by other parties, which can further boost model quality improvements. 

\textbf{Unified engine for detection and RCA} Trace graph contains rich information about the system at a given time. With the help of trace data, incident detection and root cause localization can be done within one step, instead of the current two consecutive steps. Existing work has demonstrated that by simply examining the constructed graph, the detection model can reveal sufficient information to locate the root causes \cite{Zhang2022}.

\textbf{Unified models for multimodal telemetry data} Trace data analysis brings the opportunities for researchers to create a holistic view of multiple telemetry data modality since traces can be converted into text sequence data and time-series data. The learnings can be extended to include logs or metrics from different sources. Eventually we can expect unified learning models that can consume multimodal telemetry data for incident detection and RCA.

\textbf{Online Learning} Modern systems are dynamic and ever-changing. Current two-step solution relies on offline model training and online serving or inference. Any system evolution between two offline training cycles could cause potential issues and damage model performance. Thus, supporting online learning is critical to guarantee high performance in real production environments.

%% file: failure_prediction.tex
Incident Detection and Root-Cause Analysis of Incidents are more reactive measures towards mitigating the effects of any incident and improving service availability once the incident has already occurred. On the other hand, there are other proactive actions that can be taken to predict if any potential incident can happen in the immediate future and prevent it from happening. Failures in software systems are such kind of highly disruptive incidents that often start by showing symptoms of deviation from the normal routine behavior of the required system functions and typically result in failure to meet the service level agreement. Failure prediction is one such proactive task in Incident Management, whose objective is to continuously monitor the system health by analyzing the different types of system data (KPI metrics, logging and trace data) and generate early warnings to prevent failures from occurring. Consequently, in order to handle the different kinds of telemetry data sources, the task of predicting failures can be tailored to metric based and log based failure prediction. We describe these two in details in this section. 

\subsection{Metrics based Failure Prediction}
\label{sec:failure_metrics}
\input{metric_fail}

\subsection{Logs based Incident Detection}
\label{sec:failure_logs}
\input{log_fail}

%% file: metric_fail.tex
Metric data are usually fruitful in monitoring system. It is straightforward to directly leverage them to predict the occurrence of the incident in advance. As such, some proactive actions can be taken to prevent it from happening instead of reducing the time for detection. Generally, it can be formulated as the imbalanced binary classification problem if failure labels are available, and formulated as the time series forecasting problem if the normal range of monitored metrics are defined in advance. In general, failure prediction \cite{DBLP:journals/csur/SalfnerLM10} usually adopts machine learning algorithms to learn the characteristics of historical failure data, build a failure prediction model, and then deploy the model to predict the likelihood of a failure in the future.


\vspace{1em}
\textbf{Methods}

\textbf{General Failure Prediction:}
Recently, there are increasing efforts on
considering general failure incident prediction with the failure signals from  the whole monitoring system. \cite{DBLP:conf/www/ChenYLZGXDZDXLK19} collected alerting signals across the whole system and discovered the dependence relationships among alerting signals, then the gradient boosting tree based model was adopted to learn failure patterns. \cite{DBLP:conf/sigsoft/ZhaoCWPWWZFNZSP20} proposed an effective feature engineering process to deal with complex alert data. It used multi-instance learning and handle noisy alerts, and interpretable analysis to generate an interpretable prediction result to facilitate the understanding and handling of incidents.

\textbf{Specific Type Failure Prediction:}
In contrast, some works In contrast, \cite{DBLP:conf/www/ChenYLZGXDZDXLK19} and \cite{DBLP:conf/sigsoft/ZhaoCWPWWZFNZSP20} aim to
proactively predict various specific types of failures.
\cite{DBLP:journals/pomacs/ZhangLMLBYLPXZC18} extracted statistical and textual features from historical switch logs and applied random forest to predict switch failures in data center networks. \cite{DBLP:conf/sigsoft/LinHDZSXLLWYCZ18} collected data from SMART \cite{DBLP:conf/fast/PinheiroWB07} and system-level signals, and proposed a hybrid of LSTM and random forest model for node failure prediction in cloud service system. 
\cite{DBLP:conf/usenix/XuSYZLDLJZLCZ18} developed a disk error prediction method via a cost-sensitive ranking models. These methods target at the specific type of failure prediction, and thus are limited in practice.

\vspace{1em}
\textbf{Challenges and Future Trends}

While conventional supervised learning for classification or regression problems can be used to handle failure prediction, it needs to overcome the following main challenges. First, datasets are usually very imbalanced due to the limited number of failure cases. This poses a significant challenge to the prediction model to achieve high precision and high recall simultaneously. Second, the raw signals are usually noisy, not all information before incident is helpful. How to extract omen features/patterns and filter out noises are critical to the prediction performance. Third, it is common for a typical system to generate a large volume of signals per minute, leading to the challenge to update prediction model in the streaming way and handle the large-scale data with limited computation resources. Fourth, post-processing of failure prediction is very important for failure management system to improve availability. For example, providing interpretable failure prediction can facilitate engineers to take appropriate action for it.

%% file: log_fail.tex
Like Incident Detection and Root Cause Analysis, Failure Prediction is also an extremely complex task, especially in enterprise level systems which comprise of many distributed but inter-connected components, services and micro-services interacting with each other asynchronously. One of the main complexities of the task is to be able to do early detection of signals alluding towards a major disruption, even while the system might be showing only slight or manageable deviations from its usual behavior. Because of this nature of the problem, often monitoring the KPI metrics alone may not suffice for early detection, as many of these metrics might register a late reaction to a developing issue or may not be fine-grained enough to capture the early signals of an incident. System and software logs, on the other hand, being an all-pervasive part of systems data continuously capture rich and very detailed runtime information that are often pertinent to detecting possible future failures.  

Thus various proactive log based analysis have been applied in different industrial applications as a continuous monitoring task and have proved to be quite effective for a more fine-grained failure prediction and localizing the source of the potential failure. It involves analyzing the sequences of events in the log data and possibly even correlating them with other data sources like metrics in order to detect anomalous event patterns that indicate towards a developing incident. This is typically achieved in literature by employing supervised or semi-supervised machine learning models to predict future failure likelihood by learning and modeling the characteristics of historical failure data. In some cases these models can also be additionally powered by domain knowledge about the intricate relationships between the systems. While this task has not been explored as popularly as Log Anomaly Detection and Root Cause Analysis and there are fewer public datasets and benchmark data, software and systems maintainance logging data still plays a very important role in predicting potential future failures. In literature, generally the failure prediction task over log data has been employed in broadly two types of systems - homogenous and heterogenous. 

\vspace{1em}
\textbf{Failure Prediction in Homogenous Systems} 

In homogenous systems, like high-performance computing systems or large-scale supercomputers, this entails prediction of independent failures, where most systems leverage sequential information to predict failure of a single component. 

\textbf{Time-Series Modeling}: Amongst homogenous systems, \cite{10.1145/956750.956799,9566190} extract system health indicating features from structured logs and modeled this as time series based anomaly forecasting problem. Similarly \cite{8049014} extracts specific patterns during critical events through feature engineering and build a supervised binary classifier to predict failures. \cite{10.1145/3179405} converts unstructured logs into templates through parsing and apply feature extraction and time-series modeling to predict surge, frequency and seasonality patterns of anomalies. 

\textbf{Supervised Classifiers}  Some of the older works predict failures in a supervised classification setting using traditional machine learning models like support vector machines, nearest-neighbor or rule-based classifiers \cite{DBLP:journals/ese/RussoSP15,4470294,10.1016/j.jss.2012.06.025}, or ensemble of classifiers \cite{4470294} or hidden semi-markov model based classifier \cite{4365693}  over features handcrafted from log event sequences or over random indexing based log encoding while \cite{10.1145/3208040.3208051,9006011} uses deep recurrent neural models like LSTM over semantic representations of logs. \cite{5270289} predict and diagnose failures through first failure identification and causality based filtering to combine correlated events for filtering through association rule-mining method.  

\vspace{1em}
\textbf{Failure Prediction in Heterogenous Systems} 

In heterogenous systems, like large-scale cloud services, especially in distributed micro-service environment, outages can be caused by heterogenous components. Most popular methods utilize knowledge about the relationship and dependency between the system components, in order to predict failures. Amongst such systems, \cite{10.1145/3308558.3313501} constructed a Bayesian network to identify conditional dependence between alerting signals extracted from system logs and past outages in offline setting and used gradient boosting trees to predict future outages in the online setting. \cite{10.1145/3236024.3236060} uses a ranking model combining temporal features from LSTM hidden states and spatial features from Random Forest to rank relationships between failure indicating alerts and outages. \cite{10.1145/3338906.3338961} trains trace-level and micro-service level prediction models over handcrafted features extracted from trace logs to detect three common types of micro-service failures.

%% file: root_cause_analysis.tex
Root-cause Analysis (RCA) is the process to conduct a series of actions to discover the root causes of an incident. RCA in DevOps focuses on building the standard process workflow to handle incidents more systematically. Without AI, RCA is more about creating rules that any DevOps member can follow to solve repeated incidents. However, it is not scalable to create separate rules and process workflow for each type of repeated incident when the systems are large and complex. AI models are capable to process high volume of input data and learn representations from existing incidents and how they are handled, without humans to define every single details of the workflow. Thus, AI-based RCA has huge potential to reform how root cause can be discovered. 

In this section, we discuss a series of AI-based RCA topics, separeted by the input data modality: metric-based, log-based, trace-based and multimodal RCA.

\subsection{Metric-based RCA}
\label{sec:rca_metrics}
\input{metric_rca}

\subsection{Log-based RCA}
\label{sec:rca_logs}
\input{log_rca}

\subsection{Trace-based and Multimodal RCA}
\label{sec:trace_rca}
\input{trace_rca}

%% file: metric_rca.tex
\vspace{1em}
\textbf{Problem Definition}

With the rapidly growing adoption of microservices architectures, multi-service applications become the standard paradigm in real-world IT applications. A multi-service application usually contains hundreds of interacting services, making it harder to detect service failures and identify the root causes. Root cause analysis (RCA) methods leverage the KPI metrics monitored on those services to determine the root causes when a system failure is detected, helping engineers and SREs in the troubleshooting process\footnote{A good survey for anomaly detection and RCA in cloud applications \cite{10.1145/3501297}}. The key idea behind RCA with KPI metrics is to analyze the relationships or dependencies between these metrics and then utilize these relationships to identify root causes when an anomaly occurs. Typically, there are two types of approaches: 1) identifying the anomalous metrics in parallel with the observed anomaly via metric data analysis, and 2) discovering a topology/causal graph that represent the causal relationships between the services and then identifying root causes based on it.

\vspace{1em}
\textbf{Metric Data Analysis}


When an anomaly is detected in a multi-service application, the services whose KPI metrics are anomalous can possibly be the root causes. The first approach directly analyzes these KPI metrics to determine root causes based on the assumption that significant changes in one or multiple KPI metrics happen when an anomaly occurs. Therefore, the key is to identify whether a KPI metric has pattern or magnitude changes in a look-back window or snapshot of a given size at the anomalous timestamp. 

Nguyen \etal \cite{10.1145/2038633.2038634,6681572} propose two similar RCA methods by analyzing low-level system metrics, e.g., CPU, memory and network statistics. Both methods first detect abnormal behaviors for each component via a change point detection algorithm when a performance anomaly is detected, and then determine the root causes based on the propagation patterns obtained by sorting all critical change points in a chronological order. Because a real-world multi-service application usually have hundreds of KPI metrics, the change point detection algorithm must be efficient and robust. \cite{10.1145/2038633.2038634} provides an algorithm by combining cumulative sum charts and bootstrapping to detect change points. To identify the critical change point from the change points discovered by this algorithm, they use a separation level metric to measure the change magnitude for each change point and extract the critical change point whose separation level value is an outlier. Since the earliest anomalies may have propagated from their corresponding services to other services, the root causes are then determined by sorting the critical change points in a chronological order. To further improve root cause pinpointing accuracy, \cite{6681572} develops a new fault localization method by considering both propagation patterns and service component dependencies.

Instead of change point detection, Shan \etal \cite{10.1145/3308558.3313653} developed a low-cost RCA method called $\epsilon$-Diagnosis to detect root causes of small-window long-tail latency for web services. $\epsilon$-Diagnosis assumes that the root cause metrics of an abnormal service have significantly changes between the abnormal and normal periods. It applies the two-sample test algorithm and $\epsilon$-statistics for measuring similarity of time series to identify root causes. In the two-sample test, one sample (normal sample) is drawn from the snapshot during the normal period while the other sample (anomaly sample) is drawn during the anomalous period. If the difference between the anomaly sample and the normal sample are statistically significant, the corresponding metrics of the samples are potential root causes.

\vspace{1em}
\textbf{Topology or Causal Graph-based Analysis}

The advantage of metric data analysis methods is the ability of handling millions of metrics. But most of them don't consider the dependencies between services in an application. The second type of RCA approaches leverages such dependencies, which usually involves two steps, i.e., constructing topology/causal graphs given the KPI metrics and domain knowledge, and extracting anomalous subgraphs or paths given the observed anomalies. Such graphs can either be reconstructed from the topology (domain knowledge) of a certain application (\cite{10.1145/3135974.3135977,9110353,BRANDON2020110432,8972825}) or automatically estimated from the metrics via causal discovery techniques (\cite{8411065,8367054,7563819,9213058,Lin2018MicroscopePP,8818432,10.1145/3366423.3380111}).
To identify the root causes of the observed anomalies, random walk (e.g., \cite{10.1145/2465529.2465753,9213058,8411065}), page-rank (e.g., \cite{9110353}) or other techniques can be applied over the discovered topology/causal graphs.

When the service graphs (the relationships between the services) or the call graphs (the communications among the services) are available, the topology graph of a multi-service application can be reconstructed automatically, e.g., \cite{10.1145/3135974.3135977,9110353}. But such domain knowledge is usually unavailable or partially available especially when investigating the relationships between the KPI metrics instead of API calls. Therefore, given the observed metrics, causal discovery techniques, e.g., \cite{Spirtes1991,zbMATH01835994,zbMATH02111315} play a significant role in constructing the causal graph describing the causal relationships between these metrics. The most popular causal discovery algorithm applied in RCA is the well-known PC-algorithm \cite{Spirtes1991} due to its simplicity and explainability. It starts from a complete undirected graph and eliminates edges between the metrics via conditional independence test. The orientations of the edges are then determined by finding V-structures followed by orientation propagation. Some variants of the PC-algorithm \cite{doi:10.1126/sciadv.aau4996,Runge2020DiscoveringCA,NEURIPS2020_94e70705} can also be applied based on different data properties.

Given the discovered causal graph, the possible root causes of the observed anomalies can be determined by random walk. A random walk on a graph is a random process that begins at some node, and randomly moves to another node at each time step. The probability of moving from one node to another is defined in the the transition probability matrix. Random walk for RCA is based on the assumption that a metric that is more correlated with the anomalous KPI metrics is more likely to be the root cause. Each random walk starts from one anomalous node corresponding to an anomalous metric, then the nodes visited the most frequently are the most likely to be the root causes. The key of random walk approaches is to determine the transition probability matrix. Typically, there are three steps for computing the transition probability matrix, i.e., forward step (probability of walking from a node to one of its parents), backward step (probability of walking from a node to one of its children) and self step (probability of staying in the current node). For example, \cite{8411065,8818432,10.1145/3366423.3380111,9110353} computes these probabilities based on the correlation of each metric with the detected anomalous metrics during the anomaly period. But correlation based random walk may not accurately localize root cause \cite{9213058}. Therefore, \cite{9213058} proposes to use the partial correlations instead of correlations to compute the transition probabilities, which can remove the effect of the confounders of two metrics. 

Besides random walk, other causal graph analysis techniques can also be applied. For example, \cite{Lin2018MicroscopePP,7563819} find root causes for the observed anomalies by recursively visiting all the metrics that are affected by the anomalies, e.g., if the parents of an affected metric are not affected by the anomalies, this metric is considered a possible root cause. \cite{app10062166} adopts a search algorithm based on a breadth-first search (BFS) algorithm to find root causes. The search starts from one anomalous KPI metric and extracts all possible paths outgoing from this metric in the causal graph. These paths are then sorted based on the path length and the sum of the weights associated to the edges in the path. The last nodes in the top paths are considered as the root causes. \cite{Budhathoki2022} considers counterfactuals for root cause analysis based on the causal graph, i.e., given a functional causal model, it finds the root cause of a detected anomaly by computing the contribution of each noise term to the anomaly score, where the contributions are symmetrized using the concept of Shapley values.

\vspace{1em}
\textbf{Limitations}

\textbf{Data Issues} For a multi-service application with hundreds of KPI metrics monitored on each service, it is very challenging to determine which metrics are crucial for identifying root causes. The collected data usually doesn't describe the whole picture of the system architecture, e.g., missing some important metrics. These missing metrics may be the causal parents of other metrics, which violates the assumption of PC algorithms that no latent confounders exist. Besides, due to noises, non-stationarity and nonlinear relationships in real-world KPI metrics, recovering accurate causal graphs becomes even harder.

\textbf{Lack of Domain Knowledge} The domain knowledge about the monitored application, e.g., service graphs and call graphs, is valuable to improve RCA performance. But for a complex multi-service application, even developers may not fully understand the meanings or the relationships of all the monitored metrics. Therefore, the domain knowledge provided by experts is usually partially known, and sometimes conflicts with the knowledge discovered from the observed data.

\textbf{Causal Discovery Issues} The RCA methods based on causal graph analysis leverage causal discovery techniques to recover the causal relationships between KPI metrics. All these techniques have certain assumptions on data properties which may not be satisfied with real-world data, so the discovered causal graph always contains errors, e.g., incorrect links or orientations. In recent years, many causal discovery methods have been proposed with different assumptions and characteristics, so that it is difficult to choose the most suitable one given the observed data. 

\textbf{Human in the Loop} After DevOps or SRE teams receive the root causes identified by a certain RCA method, they will do further analysis and provide feedback about whether these root causes make sense. Most RCA methods cannot leverage such feedback to improve RCA performance, or provide explanations why the identified root causes are incorrect.

\textbf{Lack of Benchmarks} Different from incident detection problems, we lack benchmarks to evaluate RCA performance, e.g., few public datasets with groundtruth root causes are available, and most previous works use private internal datasets for evaluation. Although some multi-service application demos/simulators can be utilized to generate synthetic datasets for RCA evaluation, the complexity of these demo applications is much lower than real-world applications, so that such evaluation may not reflect the real performance in practice. The lack of public real-world benchmarks hampers the development of new RCA approaches.

\vspace{1em}
\textbf{Future Trends}

\textbf{RCA Benchmarks} Benchmarks for evaluating the performance of RCA methods are crucial for both real-world applications and academic research. The benchmarks can either be a collection of real-world datasets with groundtruth root causes or some simulators whose architectures are close to real-world applications. Constructing such large-scale real-world benchmarks is essential for boosting novel ideas or approaches in RCA.

\textbf{Combining Causal Discovery and Domain Knowledge} The domain knowledge provided by experts are valuable to improve causal discovery accuracy, e.g., providing required or forbidden causal links between metrics. But sometimes such domain knowledge introduces more issues when recovering causal graphs, e.g., conflicts with data properties or conditional independence tests, introducing cycles in the graph. How to combine causal discovery and expert domain knowledge in a principled manner is an interesting research topic. 

\textbf{Putting Human in the Loop} Integrating human interactions into RCA approaches is important for real-world applications. For instance, the causal graph can be built in an iterative way, i.e., an initial causal graph is reconstructed by a certain causal discovery algorithm, and then users examine this graph and provide domain knowledge constraints (e.g., which relationships are incorrect or missing) for the algorithm to revise the graph. The RCA reports with detailed analysis about incidents created by DevOps or SRE teams are valuable to improve RCA performance. How to utilize these reports to improve RCA performance is another importance research topic.

%% file: log_rca.tex
\textbf{Problem Definition}

Triaging and root cause analysis is one of the most complex and critical phases in the Incident Management life cycle. Given the nature of the problem which is to investigate into the origin or the root cause of an incident, simply analyzing the end KPI metrics often do not suffice. Especially in a micro-service application setting or distributed cloud environment with hundreds of services interacting with each other, RCA and failure diagnosis is particularly challenging. In order to localize the root cause in such complex environments, engineers, SREs and service owners typically need to investigate into core system data. Logs are one such ubiquitous forms of systems data containing rich runtime information. Hence one of the ultimate objectives of log analysis tasks is to enable triaging of incident and localization of root cause to diagnose faults and failures. 

Starting with heterogenous log data from different sources and microservices in the system, typical log-based aiops workflows first have a layer of log processing and analysis, involving log parsing, clustering, summarization and anomaly detection. The log analysis and anomaly detection can then cater to a causal inference layer that analyses the relationships and dependencies between log events and possibly detected anomalous events. These signals extracted from logs within or across different services can be further correlated with other observability data like metrics, traces etc in order to detect the root cause of an incident. Typically this involves constructing a causal graph or mining a knowledge graph over the log events and correlating them with the KPI metrics or with other forms of system data like traces or service call graphs. Through these, the objective is to analyze the relationships and dependencies between them in order to eventually identify the possible root causes of an anomaly. Unlike the more concrete problems like log anomaly detection, log based root cause analysis is a much more open-ended task. Subsequently most of the literature on log based RCA has been focused on industrial applications deployed in real-world and evaluated with internal benchmark data gathered from in-house domain experts. 

\vspace{1em}
\textbf{Typical types of Log RCA methods} 

In literature, the task of log based root cause analysis have been explored through various kinds of approaches. While some of the works build a knowledge graph and knowledge and leverage data mining based solutions, others follow fundamental principles from Causal Machine learning or and causal knowledge mining. Other than these, there are also log based RCA systems using traditional machine learning models which use feature engineering or correlational analysis or supervised classifier to detect the root cause.  

\vspace{0.5em}
\textbf{Handcrafted features based methods:} \cite{8029786} uses handcrafted feature engineering and probabilistic estimation of specific types of root causes tailored for Spark logs.  \cite{10.1145/3392149} uses frequent item-set mining and association rule mining on feature groups for structured logs.

\vspace{0.5em}
\textbf{Correlation based Methods:} \cite{9283922,10.1145/2623330.2623374} localizes root cause based on correlation analysis using mutual information between anomaly scores obtained from logs and monitored metrics. Similarly \cite{5713159} use PCA, ICA based correlation analysis to capture relationships between logs and consequent failures. \cite{DBLP:conf/sosp/XuHFPJ09,8952396} uses PCA to detect abnormal system call sequences which it maps to application functions through frequent pattern mining.\cite{7840733} uses LSTM based sequential modeling of log templates identified through pattern matching over clusters of similar logs, in order to predict failures. 

\vspace{0.5em}
\textbf{Supervised Classifier based Methods:} \cite{8814553} does automated detection of exception logs and comparison of new error patterns with normal cloud behaviours on OpenStack by learning supervised classifiers over statistical and neural representations of historical failure logs. \cite{6410318} employs statistical technique on the data distribution to identify the fine-grained category of a performance problem and fast matrix recovery RPCA to identify the root cause. \cite{10.1109/ICSE.2017.71,8812113} uses KNN or its supervised versions to identify loglines that led to a failure. 

\vspace{0.5em}
\textbf{Knowledge Mining based Methods:} \cite{10.1145/3502223.3502250,10.1007/978-3-030-77385-4_38} takes a different approach of summarizing log events into an entity-relation knowledge graph by extracting custom entities and relationships from log lines and mining temporal and procedural dependencies between them from the overall log dump. While this gives a more structured representation of the log summary, it is also an intuitive way of aggregating knowledge from logs, it is also a way to bridge the knowledge gap developer community who creates the log data and the site reliability engineers who typically consume the log data when investigating incidents. However, eventually the end goal of constructing this knowledge graph representation of logs is to facilitate RCA. While these works do provide use-cases like case-studies on RCA for this vision, but they leave ample scope of research towards a more concrete usage of this kind of knowledge mining in RCA. 

\vspace{0.5em}
\textbf{Knowledge Graph based Methods:} Amongst knowledge graph based methods, \cite{10.1145/3377813.3381353} diagnoses and triages performance failure issues in an online fashion by continuously building a knowledge base out of rules extracted from a random forest constructed over log data using heuristics and domain knowledge. \cite{BRANDON2020110432} constructs a system graph from the combination of KPI metrics and log data. Based on the detected anomalies from these data sources, it extracts anomalous subgraphs from it and compares them with the normal system graph to detect the root cause. Other works mine normal log patterns \cite{7484164} or time-weighted control flow graphs \cite{8030620} from normal executions and on estimates divergences from them to executions during ongoing failures to suggest root causes. \cite{6606586,7484300,10.5555/2228298.2228334} mines execution sequences or  user actions \cite{8647957} either from normal and manually injected failures or from good or bad performing systems, in a knowledge base and utilizes the assumption that similar faults generate similar failures to match and diagnose type of failure. Most of these knowledge based approaches incrementally expand their knowledge or rules to cater to newer incident types over time. 

\vspace{0.5em}
\textbf{Causal Graph based Methods: }\cite{10.1007/978-3-030-76352-7_17} uses a multivariate time-series modeling over logs by representing them as error event count. This work then infers its causal relationship with KPI error rate using a pagerank style centrality detection in order to identify the top root causes. \cite{app10062166} constructs a knowledge graph over operation and maintenance entities extracted from logs, metrics, traces and system dependency graphs and mines causal relations using PC algorithm to detect root causes of incidents. \cite{10.1145/3459637.3481903} uses a Knowledge informed Hierarchical Bayesian Network over features extracted from metric and log based anomaly detection to infer the root causes. \cite{10.1109/ASE51524.2021.9678708} constructs dynamic causality graph over events extracted from logs, metrics and service dependency graphs. \cite{6968768} similarly constructs a causal dependency graph over log events by clustering and mining similar events and use it to infer the process in which the failure occurs.  

Also, on a related domain of network analysis, \cite{7987263,9012718,9529498} mines causes of network events through causal analysis on network logs by modeling the parsed log template counts as a multivariate time series. \cite{10.1109/ICSE-SEIP52600.2021.00043,9213058} use causality inference on KPI metrics and service call graphs to localize root causes in microservice systems and one of the future research directions is to also incorporate unstructured logs to such causal analysis.

\vspace{1em}

\textbf{Challenges \& Future Trends}
\vspace{0.5em}
\textbf{Collecting supervision labels} Being a complex and open-ended task, it is challenging and requires a lot of domain expertise and manual effort to collect supervision labels for root cause analysis. While a small scale supervision can still be availed for evaluation purposes, reaching the scale required for training these models is simply not practical. At the same time, because of the complex nature of the problem, completely unsupervised models often perform quite poorly. 
\vspace{0.5em}
\textbf{Data quality:} The workflow of RCA over heterogeneous unstructured log data typically involves various different analysis layers, preprocessing, parsing, partitioning and anomaly detection. This results in compounding and cascading of errors (both labeling errors as well as model prediction errors) from these components, needing the noisy data to be handled in the RCA task. In addition to this, the extremely challenging nature of RCA labeling task further increases the possibility of noisy data.
\vspace{0.5em}
\textbf{Imbalanced class problem:} RCA on huge voluminous logs poses an additional problem of extreme class imbalance - where out of millions of log lines or log templates, a very sparse few instances might be related to the true root cause. 
\vspace{0.5em}
\textbf{Generalizability of models:} Most of the existing literature on RCA tailors their approach very specifically towards their own application and cannot be easily adopted even by other similar systems. This alludes towards need for more generalizable architectures for modeling the RCA task which in turn needs more robust generalizable log analysis models that can handle hetergenous kinds of log data coming from different systems.  
\vspace{0.5em}
\textbf{Continual learning framework:} One of the challenging aspects of RCA in the distributed cloud setting is the agile environment, leading to new kinds of incidents and evolving causation factors. This kind of non-stationary learning setting poses non-trivial challenges for RCA but is indeed a crucial aspect of all practical industrial applications.
\vspace{0.5em}
\textbf{Human-in-the-loop framework:} While neither completely supervised or unsupervised settings is practical for this task, there is need for supporting human-in-the-loop framework which can incorporate feedbacks from domain experts to improve the system, especially in the agile settings where causation factors can evolve over time.
\vspace{0.5em}
\textbf{Realistic public benchmarks:} Majority of the literature in this area is focused on industrial applications with in-house evaluation setting. In some cases, they curate their internal testbed by injecting failures or faults or anomalies in their internal simulation environment (for e.g. injecting CPU, memory, network and Disk anomalies in Spark platforms) or in  popular testing settings (like Grid5000 testbed or open-source microservice applications based on online shopping platform or train ticket booking or open source cloud operating system OpenStack). Other works evaluate by deploying their solution in real-world setting in their in-house cloud-native application, for e.g. on IBM Bluemix platform, or for Facebook applications or over hundreds of real production services at big data cloud computing platforms like Alibaba or thousands of services at e-commerce enterprises like eBay. One of the striking limitations in this regard is the lack of any reproducible open-source public benchmark for evaluating log based RCA in practical industrial settings.  This can hinder more open ended research and fair evaluation of new models for tackling this challenging task.

%% file: trace_rca.tex
\vspace{1em}
\textbf{Problem Definition.} Ideally, RCA for a complex system needs to leverage all kind of available data, including machine generated telemetry data and human activity records, to find potential root causes of an issue. In this section we discuss trace-based RCA together with multi-modal RCA. We also include studies about RCA based on human records such as incident reports. Ultimately, the RCA engine should aim to process any data types and discover the right root causes.

\vspace{1em}
\textbf{RCA on Trace Data}

In previous section (Section \ref{sec:trace_ad}) we discussed trace can be treated as multimodal data for anomaly detection. Similar to trace anomaly detection, trace root cause analysis also leverages the topological structure of the service map. Instead of detecting abnormal traces or paths, trace RCA usually started after issues were detected. Trace RCA techniques help ease troubleshooting processes of engineers and SREs. And trace RCA can be triggered in a more ad-hoc way instead of running continuously. This differentiates the potential techniques to be adopted from trace anomaly detection. 

\textbf{Trace Entity Graph.} From the technical point of view, trace RCA and trace anomaly detection share similar perspectives. To our best knowledge, there are not too many existing works talking about trace RCA alone. Instead, trace RCA serves as an additional feature or side benefit for trace anomaly detection in either empirical approaches \cite{Arnold2007} \cite{Li2021} or deep learning approaches \cite{Zhang2022} \cite{Zhou2019}. In trace anomaly detection, the constructed trace entity graph (TEG) after offline training provides a clean relationship between each component in the application systems. Thus, besides anomaly detection, \cite{cai2019real} implemented a real-time RCA algorithm that discovers the deepest root of the issues via relative importance analysis after comparing the current abnormal trace pattern with normal trace patterns. Their experiment in the production environment demonstrated this RCA algorithm can achieve higher precision and recall compared to naive fixed threshold methods. The effectiveness of leverage trace entity graph for root cause analysis is also proven in deep learning based trace anomaly detection approaches. Liu \textit{et al.} \cite{Liu2020} proposed a multimodal LSTM model for trace anomaly detection. Then the RCA algorithm can check every anomalous trace with the model training traces and discover root cause by localizing the next called microservice which is not in the normal call paths. This algorithm performs well for both synthetic dataset and production datasets of four large production services, according to the evaluation of this work. 

\textbf{Online Learning.} An alternative approach is using data mining and statistical learning techniques to run dynamic analysis without constructing the offline trace graph. Traditional trace management systems usually provides basic analytical capabilities to diagnose issues and discover root causes \cite{Sigelman2010}. Such analysis can be performed online without costly model training process. Chen \etal proposed Pinpoint \cite{Chen2002}, a framework for root cause analysis that using coarse-grained tagging data of real client requests at real-time when these requests traverse through the system, with data mining techniques. Pinpoint discovers the correlation between success / failure status of these requests and fault components. The entire approach processes the traces on-the-fly and does not leverage any static dependency graph models. Another related area is using trouble-shooting guide data, where \cite{10.1145/3368089.3417054} recommends troubleshooting guide based on semantic similarity with incident description while \cite{DBLP:journals/corr/abs-2205-13457} focuses on automation of troubleshooting guides to execution workflows, as a way to remediate the incident.

\vspace{1em}
\textbf{RCA on Incident Reports} 

Another notable direction in AIOps literature has been mining useful knowledge from domain-expert curated data (incident report, incident investigation data, bug report etc) towards enabling the final goals of root cause analysis and automated remediation of incidents. This is an open ended task which can serve various purposes - structuring and parsing unstructured or semi-structured data and extracting targeted information or topics from them (using topic modeling or information extraction) and mining and aggregating knowledge into a structured form. 

The end-goal of these tasks is majorly root cause analysis, while some are also focused on recommending remediation to mitigate the incident. Especially since in most cloud-based settings, there is an increasing number of incidents that occur repeatedly over time showing similar symptoms and having similar root causes. This makes mining and curating knowledge from various data sources, very crucial, in order to be consumed by data-driven AI models or by domain experts for better knowledge reuse. 

\textbf{Causality Graph.} \cite{7820614} extracts and mines causality graph from historical incident data and uses human-in-the-loop supervision and feedback to further refine the causality graph. \cite{8711092} constructs an anomaly correlation graph, FacGraph using a distributed frequent pattern mining algorithm. \cite{6903589} recommends appropriate healing actions by adapting remediations retrieved from similar historical incidents. Though the end task involves remediation recommendation, the system still needs to understand the nature of incident and root cause in order to retrieve meaningful past incidents.

\textbf{Knowledge Mining.} \cite{10.1109/ICSE-SEIP52600.2021.00031,DBLP:journals/ese/ShettyBKRN22} mines knowledge graph from  named entity and relations extracted from incident reports using LSTM based CRF models. \cite{DBLP:conf/icse/SahaH22} extracts symptoms, root causes and remediations from past incident investigations and builds a neural search and knowledge graph to facilitate a retrieval based root cause and remediation recommendation for recurring incidents. 

\vspace{1em}
\textbf{Future Trends}

\textbf{More Efficient Trace Platform.} Currently there are very limited studies in trace related topics. A fundamental challenge is about the trace platforms.There are bottlenecks in collection, storage, query and management of trace data. Traces are usually at a much larger scale than logs and metrics. How to more efficiently collect, store and retrieve trace data is very critical to the success of trace root cause analysis. 

\textbf{Online Learning.} Compared to trace anomaly detection, online learning plays a more important role for trace RCA, especially for large cloud systems. An RCA tool usually needs to analyze the evidence on the fly and correlate the most suspicious evidence to the ongoing incidents, this approach is very time sensitive. For example, we know trace entity graph (TEG) can achieve accurate trace RCA but the preassumpiton is the TEG is reflecting the current status of the system. If offline training is the only way to get TEG, the performance of such approaches in real-world production environments is always questionable. Thus, using online learning to obtain the TEG is a much better way to guarantee high performance in this situation.

\textbf{Causality Graphs on Multimodal Telemetries.} The most precious information conveyed by trace data is the complex topological order of large systems. Without traces, causal analysis for system operations relies on temporal and geometrical correlations to infer causal relationships, and practically very few existing causal inference can be adopted in real-world systems. However, with traces, it is very convenient to obtain the ground truth of how requests flow through the entire system. Thus, we believe higher quality causal graphs will be much easier achievable if it can be learned by multimodel telemetry data. 

\textbf{Complete Knowledge Graph of Systems.} Currently knowledge mining has been tried for single data type. However, to reflect the full picture of a complex system, the AI models need to mining knowledge from any kind of data types, including metrics, logs, traces, incident reports and other system activity records, then construct a knowledge graph with complete system information.

%% file: automated_actions.tex
While both incident detection and RCA capabilities of AIOps help provide information about ongoing issues, taking the right actions is the step that solve the problems. Without automation to take actions, human operators will still be needed in every single ops task. Thus, automated actions is critical to build fully-automated end-to-end AIOps systems. Automated actions contributes to both short-term actions and longer-term actions: 1) \textit{short-term remediation}: immediate actions to quickly remediate the issue, including server rebooting, live migration, automated scaling, etc.; and 2) \textit{longer-term resolutions}: actions or guidance for tasks such as code bug fixing, software updating, hard build-out and resource allocation optimization. In this section, we discuss three common types of automated actions: automated remediation, auto-scaling and resource management.

\subsection{Automated Remediation}
\label{sec:auto_remediation}
\input{auto_remediation}

\subsection{Auto-scaling}
\label{sec:auto-scaling}
\input{autoscaling}

\subsection{Resource Management}
\label{sec:resource_management}
\input{resource_management}

%% file: auto_remediation.tex
\vspace{1em}
\textbf{Problem Definition}

Besides continuously monitoring the IT infrastructure, detecting issues and discovering root causes, remediating issues with minimum, or even no human intervention, is the path towards the next generation of fully automated AIOps. Automated issue remediation (Auto-remediation) is taking a series of actions to resolve issues by leveraging known information, existing workflows and domain knowledge. Auto-remediation is a concept already adopted in many IT operation scenarios, including cloud computing, edge computing, SaaS, etc. 

Traditional auto-remediation processes are based on a variety of well-defined policies and rules to get which workflows to use for a given issue. While machine learning driven auto-remediation means utilizing machine learning models to decide the best action workflows to mitigate or resolve the issue. ML based auto-remediation is exceptionally useful in large scale cloud systems or edge-computing systems where it’s impossible to manually create workflows for all issue categories. 

\vspace{1em}
\textbf{Existing Work}

End-to-end auto-remediation solutions usually contain three main components: anomaly or issue detection, root cause analysis and remediation engine \cite{Becker2020}. This means successful auto-remediation solutions highly rely on the quality of anomaly detection and root cause analysis, which we’ve already discussed in the above sections. Besides, the remediation engine should be able to learn from the analysis results, make decisions and execute. 

\textbf{Knowledge learning. } 
The knowledge here refers to a variety of categories. Anomaly detection and root cause analysis for this specific issue contributes to a majority of the learnable knowledge \cite{Becker2020}. Remediation engine uses these information to locate and categorize the issue. Besides, the human activity records (such as tickets, bug fixing logs) of past issues are also significant for the remediation to learn the full picture of how issues were handled in history. In Sections  \ref{sec:rca_metrics} \ref{sec:rca_logs} \ref{sec:trace_rca} we discussed about mining knowledge graphs from system metrics, logs and human-in-the-loop records. A high quality knowledge graph which clearly describes the relationship of system components. 

\textbf{Decision making and execution. }
Levy \textit{et al.} \cite{Levy2020} proposed Narya, a system to handle failure remediation for running virtual machines in cloud systems. For a given issue where the host is predicted to fail, the remediation engine needs to decide what is the best action to take from a few options such as live migration, soft reboot, service healing, etc. The decision on which actions to take are made via A/B testing and reinforcement learning. With adopting machine learning in their remediation engine, they see significant virtual machine interruption savings compared to the previous static strategies.

\vspace{1em}
\textbf{Future Trends}

Auto-remediation research and development is still in very early stages. The existing work mainly focuses on an intermediate step such as constructing a causal graph for a given scenario, or an end-to-end auto-remediation solution for very specific use cases such as virtual machine interruptions. Below are a few topics that can significantly improve the quality of auto-remediation systems.

\textbf{System Integration} Now there is still no unified platform that can perform all the issue analysis, learn the context knowledge, make decisions and execute the actions. 

\textbf{Learn to generate and update knowledge graphs} Quality of auto-remediation decision making strongly depends on domain knowledge. Currently humans collect most of the domain knowledge. In the future, it is valuable to explore approaches that learn and maintain knowledge graphs of the systems in a more reliable way.

\textbf{AI driven decision making and execution} Currently most of the decision making and action execution are rule-based or statistical learning based. With more powerful AI techniques, the remediation engine can then consume rich information and make more complex decisions.

%% file: autoscaling.tex
\vspace{1em}
\textbf{Problem Definition}

The cloud native technologies are becoming the de facto standard for building scalable applications in public or private clouds, enabling loosely coupled systems that are resilient, manageable, and observable\footnote{https://github.com/cncf/foundation/blob/main/charter.md}. The cloud systems such as GCP and AWS provide users on-demand resources including CPU, storage, memory and databases. Users needs to specify a limit of these resources to provision for the workloads of their applications. If a service in an application exceeds the limit of a particular resource, end-users will experience request delays or timeouts, so that system operators will request a larger limit of this resource to avoid degraded performance. But if hundreds of services are running, such large limit results in massive resource wastage. Auto-scaling aims to resolve this issue without human intervention, which enables dynamic provisioning of resources to applications based on workload behavior patterns to minimize resource wastage without loss of quality of service (QoS) to end-users.

Auto-scaling approaches can be categorized into two types: reactive auto-scaling and proactive (or predictive) auto-scaling. Reactive auto-scaling monitors the services in a application, and brings them up and down in reaction to changes in workloads. 

\textbf{Reactive auto-scaling}. Reactive auto-scaling is very effective and supported by most cloud platforms. But it has one potential disadvantage, i.e., it won't scale up resources until workloads increase so that there is a short period in which more capacity is not yet available but workloads becomes higher. Therefore, end-users can experience response delays in this short period. Proactive auto-scaling aims to solve this problem by predicting future workloads based on historical data. In this paper, we mainly discuss proactive auto-scaling algorithms based on machine learning.

\textbf{Proactive Auto-scaling.} Typically, proactive auto-scaling involves three steps, i.e., predicting workloads, estimating capacities and scaling out. Machine learning techniques are usually applied to predict future workloads and estimate the suitable capacities for the monitored services, and then adjustments can be done accordingly to avoid degraded performance.

One type of proactive auto-scaling approaches applies regression models (e.g., ARIMA \cite{James1994}, SARIMA \cite{Hyndman2018}, MLP, LSTM \cite{HochSchm97}). Given the historical metrics of a monitored service, this type of approaches trains a particular regression model to learn the workload behavior patterns. For example, \cite{6881647} investigated the ARIMA model for workload prediction and showed that the model improves efficiency in resource utilization with minimal impact in QoS. \cite{9284206} applied a time window MLP to predict phases in containers with different types of workloads and proposed a predictive vertical auto-scaling policy to resize containers. \cite{7892598} also leveraged neural networks (especially MLP) for workload prediction and compared this approach with traditional machine learning models, e.g., linear regression and K-nearest neighbors. \cite{app11093835} applied a bidirectional LSTM to predict the number of HTTP workloads and showed that Bi-LSTM works better than LSTM and ARIMA on the tested use cases. These approaches require accurate forecasting results to avoid over- or under-allocated of resources, while it is hard to develop a robust forecasting-based approach due to the existence of noises and sudden spikes in user requests.

The other type is based on reinforcement learning (RL) that treats auto-scaling as an automatic control problem, whose goal is to learn an optimal auto-scaling policy for the best resource provision action under each observed state. \cite{GARI2021104288} presents an exhaustive survey on reinforcement learning-based auto-scaling approaches, and compares them based on a set of proposed taxonomies. This survey is very worth reading for developers or researchers who are interested in this direction. Although RL looks promising in auto-scaling, there are many issues needed to be resolved. For example, model-based methods require a perfect model of the environment and the learned policies cannot adapt to the changes in the environment, while model-free methods have very poor initial performance and slow convergence so that they will introduce high cost if they are applied in real-world cloud platforms.

%% file: resource_management.tex
\vspace{1em}
\textbf{Problem Definition}

Resource management is another important topic in cloud computing, which includes resource provisioning, allocation and scheduling, e.g., workload estimation, task scheduling, energy optimization, etc. Even small provisioning inefficiencies, such as selecting the wrong resources for a task, can affect quality of service (QoS) and thus lead to significant monetary costs. Therefore, the goal of resource management is to provision the right amount of resources for tasks to improve QoS, mitigate imbalance workloads, and avoid service level agreements violations.

Because of multiple tenants sharing storage and computation resources on cloud platforms, resource management is a difficult task that involves dynamically allocating resources and scheduling tenants' tasks. How to provision resources can be determined in a reactive manner, e.g., creating static rules manually based on domain knowledge. But similar to auto-scaling, reactive approaches result in response delays and excessive overheads. To resolve this issue, ML-based approaches for resource management have gained much attention recently.

\vspace{1em}
\textbf{ML-based Resource Management}

Many ML-based resource management approaches have been developed in recent years. Due to space limitation, we will not discuss them in details. We recommend readers who are interested in this research topic to read the following nice review papers: \cite{MUSTAFA2015186,KHAN2022103405,doi:10.1080/1206212X.2017.1416558,10.1186/s13677-017-0081-4,10.1145/3364684}. Most of these approaches apply ML techniques to forecast future resource consumption and then do resource provisioning or scheduling based on the forecasting results. For instance, \cite{10.1145/3132747.3132772} uses random forest and XGBoost to predict VM behaviors including maximum deployment sizes and workloads. \cite{Haghshenas2020} proposes a linear regression based approach to predict the resource utilization of the VMs based on their historical data, and then leverage the prediction results to reduce energy consumption. \cite{9272657} applies gradient boosting models for temperature prediction, based on which a dynamic scheduling algorithm is developed to minimize the peak temperature of hosts. \cite{10.1145/3341302.3342080} proposes a RL-based workload-specific scheduling algorithm to minimize average task completion time.

The accuracy of the ML model is the key factor that affects the efficiency of a resource management system. Applying more sophisticated traditional ML models or even deep learning models to improve prediction accuracy is a promising research direction. Besides accuracy, the time complexity of model prediction is another important factor needed to be considered. If a ML model is over-complicated, it cannot handle real-time requests of resource allocation and scheduling. How to make a trade-off between accuracy and time complexity needs to be explored further.

%% file: future.tex
\vspace{1em}
\subsection{Common AI Challenges for AIOps}

\vspace{0.5em}
We have discussed the challenges and future trends in each task sections according to how to employ AI techniques. In summary, there are some common challenges across different AIOps tasks.

\textbf{Data Quality.} For all AIOps task there are data quality issues. Most real-world AIOps data are extremely imbalanced due to the nature that incidents only occurs occasionally. Also, most of the real-world AIOps data are very noisy. Significant efforts are needed in data cleaning and pre-processing before it can be used as input to train ML models. 

\textbf{Lack of Labels.} It's extremely difficult to acquire quality labels sufficiently. We need a lot of domain experts who are very familiar with system operations to evaluate incidents, root-causes and service graphs, in order to provide high-quality labels. This is extremely time consuming and require specific expertise, which cannot be handled by general crowd sourcing approaches like Mechanical Turk.

\textbf{Non-stationarity and heterogeneity.} Systems are ever-changing. AIOps are facing non-stationary problem space. The AI models in this domain need to have mechanisms to deal with this non-stationary nature. Meanwhile, AIOps data are heterogeneous, meaning the same telemetry data can have a variety of underlying behaviors. For example, CPU utilization pattern can be totally different when the resources are used to host different applications. Thus, discovery the hidden states and handle heterogeneity is very important for AIOps solutions to succeed.

\textbf{Lack of Public Benchmarking.} Even though AIOps research communities are growing rapidly, there are still very limited number of public datasets for researchers to benchmark and evaluate their results. Operational data are highly sensitive. Existing research are done either with simulated data or enterprise production data which can hardly be shared with other groups and organizations. 

\textbf{Human-in-the-loop.} Human feedback are very important to build AIOps solutions. Currently most of the human feedback are collected in ad-hoc fashion, which is inefficient. There are lack of human-in-the-loop studies in AIOps domain to automate feedback collection and utilize the feedback to improve model performance.

\subsection{Opportunities and Future Trends}
Our literature review of existing AIOps work shows current AIOps research still focuses more on infrastructure and tooling. We see AI technologies being successfully applied in incident detection, RCA applications and some of the solutions has been adopted by large distributed systems like AWS, Alibaba cloud. While it is still in very early stages for AIOps process standardization and full automation. With these evidences, we can foresee the promising topics of AIOps in the next few years. 

\vspace{1em}
\textbf{High Quality AIOps Infrastructure and Tooling}

There are some successful AIOps platforms and tools being developed in recent years. But still there are opportunities where AI can help enhance the efficiency of IT operations. AI is also growing rapidly and new AI technologies are invented and successfully applied in other domains. The digital transformation trend also brings challenges to traditional IT operation and Devops. This creates tremendous needs for high quality AI tooling, including monitoring, detection, RCA, predictions and automations. 

\vspace{1em}
\textbf{AIOps Standardization}

While building the infrastructure and tooling, AIOps experts also better understand the full picture of the entire domain. AIOps modules can be identified and extracted from traditional processes to form its own standard. With clear goals and measures, it becomes possible to standardize AIOps systems, just as what has been done in domains like recommendation systems or NLP. With such standardization, it will be much easier to experiment a large variety of AI techniques to improve AIOps performance. 

\vspace{1em}
\textbf{Human-centric to Machine-centric AIOps}

Human-centric AIOps means human processes still play critical roles in the entire AIOps eco-systems, and AI modules help humans with better decisions and executions. While in Machine-centric mode, AIOps systems require minimum human intervention and can be in human-free state for most of its lifetime. AIOps systems continuously monitor the IT infrastructure, detecting and analysis issues, finding the right paths to drive fixes. In this stage, engineers focus primarily on development tasks rather than operations.

%% file: conclusion.tex
Digital transformation creates tremendous needs for computing resources. The trend boosts strong growth of large scale IT infrastructure, such as cloud computing, edge computing, search engines, etc. Since proposed by Gartner in 2016, AIOps is emerging rapidly and now it draws the attention from large enterprises and organizations. As the scale of IT infrastructure grows to a level where human operation cannot catch up, AIOps becomes the only promising solution to guarantee high availability of these gigantic IT infrastructures. AIOps covers different stages of software lifecycles, including development, testing, deployment and maintenance. 

Different AI techniques are now applied in AIOps applications, including anomaly detection, root-cause analysis, failure predictions, automated actions and resource management. However, the entire AIOps industry is still in a very early stage where AI only plays supporting roles to help human conducting operation workflows. We foresee the trend shifting from human-centric Operations to AI-centric Operations in the near future. During the shift, Development of AIOps techniques will also transit from build tools to create human-free end-to-end solutions. 

In this survey, we discovered that most of the current AIOps outcomes focus on detections and root cause analysis, while research work on automations is still very limited. The AI techniques used in AIOps are mainly traditional machine learning and statistical models.

%% file: appendix_terminology.tex
\textbf{DevOps:} Modern software development requires not only higher development quality but also higher operations quality. DevOps, as a set of best practices that combines the development (Dev) and operations (Ops) processes, is created to achieve high quality software development and after release management \cite{Olavsrud2021}. 

\textbf{Application Performance Monitoring (APM):} Application performance monitoring is the practice of tracking key software application performance using monitoring software and telemetry data\cite{anderson2021}. APM is used to guarantee high system availability, optimize service performance and improve user experiences. Originally APM was mostly adopted in websites, mobile apps and other similar online business applications. However, with more and more traditional softwares transforming to leverage cloud based, highly distributed systems, APM is now widely used for a larger variety of software applications and backends.

\textbf{Observability:} Observability is the ability to measure the internal states of a system by examining its outputs \cite{livens2021}. A system is “observable” if the current state can be estimated by only using the information from outputs. Observability data includes metrics, logs, traces and other system generated information.

\textbf{Cloud Intelligence:} The artificial intelligent features that improve cloud applications. 

\textbf{MLOps:} MLOps stands for machine learning operations. MLOps is the full process life cycle of deploying machine learning models to production.  

\textbf{Site Reliability Engineering (SRE):} The type of engineering that bridge the gap between software development and operations.

\textbf{Cloud Computing:} Cloud computing is a technique, and a business model, that builds highly scalable distributed computer systems and lends computing resources, e.g. hosts, platforms, apps, to tenants to generate revenue. There are three main category of cloud computing: infrastructure as a service (IaaS), platform as a service (PaaS) and software as a service (SaaS)

\textbf{IT Service Management (ITSM):} ITSM refers to all processes and activities to design, create, deliver, and support the IT services to customers.

\textbf{IT Operations Management (ITOM):} ITOM overlaps with ITSM, focusing more on the operation side of IT services and infrastructures.

%% file: appendix_tables.tex
\input{tables/metric_datasets.tex}
\input{tables/log_datasets.tex}
\input{tables/log_ad_table.tex}
\input{tables/metric_ad.tex}
\input{tables/trace_ad_table.tex}
\input{tables/metric_rca_table.tex}

%% file: tables/metric_datasets.tex
\begin{table*}[hp]
  \centering
  \caption{Table of popular public datasets for metrics observability}
\begin{tabular}{|p{0.1\textwidth}|p{0.4\textwidth}|p{0.4\textwidth}|}
\hline
\textbf{Name} & \textbf{Description} & \textbf{Tasks} \\
\hline
Azure Public Dataset & These datasets contain a representative subset of first-party Azure virtual machine workloads from a geographical region. & Workload characterization, VM Pre-provisioning, Workload prediction \\
\hline
Google Cluster Data & 30 continuous days of information from Google Borg cells. & Workload characterization, Workload prediction \\
\hline
Alibaba Cluster Trace & Cluster traces of real production servers from Alibaba Group. & Workload characterization, Workload prediction \\
\hline
MIT Supercloud Dataset & Combination of high-level data (e.g. Slurm Workload Manager scheduler data) and low-level job-specific time series data. & Workload characterization \\
\hline
Numenta Anomaly Benchmark (realAWSCloudwatch) & AWS server metrics as collected by the AmazonCloudwatch service. Example metrics include CPU Utilization, Network Bytes In, and Disk Read Bytes. & Incident detection \\
\hline
Yahoo S5 (A1) & A1 benchmark contains real Yahoo! web traffic metrics. & Incident detection \\
\hline
Server Machine Dataset & A 5-week-long dataset collected from a large Internet company containing metrics like CPU load, network usage, memory usage, etc. & Incident detection \\
\hline
KPI Anomaly Detection Dataset A & A large-scale realworld KPI anomaly detection dataset, covering various KPI patterns and anomaly patterns. This dataset is collected from five large Internet companies (Sougo, eBay, Baidu, Tencent, and Ali). & Incident detection \\
\hline
\end{tabular}%
  \label{tab:metrics_observability}%
\end{table*}%

%% file: tables/log_datasets.tex
\begin{table*}[hp]
\centering
\caption{Table of Popular Public Datasets for Log Observability}
{\scriptsize
\begin{tabular}{|p{0.08\textwidth}|p{0.2\textwidth}|p{0.1\textwidth}|p{0.07\textwidth}|p{0.08\textwidth}|p{0.05\textwidth}|p{0.1\textwidth}|p{0.1\textwidth}|} \hline
\textbf{Dataset} & \textbf{Description} & \textbf{Time-span} & \textbf{Data Size} & \textbf{\# logs} & \textbf{Anomaly Labels} & \textbf{\# Anomalies} & \textbf{\# Log Templates} \\ \hline
\multicolumn{6}{c}{\textbf{Distributed system logs}} \\ \hline 
\multirow{2}{*}{HDFS} & \multirow{2}{*}{Hadoop distributed file system log} & 38.7 hours & 1.47 GB & 11,175,629 & \cmark & 16,838(blocks) & 30 \\ \cline{3-8} 
& & N.A. & 16.06 GB & 71,118,073 & \xmark & & \\ \hline 
Hadoop & Hadoop map-reduce job log & N.A. & 48.61MB & 394,308 & \cmark & & 298 \\ \hline 
Spark & Spark job log & N.A. & 2.75GB & 33,236,604 & \xmark & & 456\\ \hline 
Zookeeper & ZooKeeper service log & 26.7 days & 9.95MB & 74,380 & \xmark & & 95\\ \hline
OpenStack & OpenStack infrastructure log & N.A. & 58.61MB & 207,820 & \cmark & 503 & 51\\ \hline 
\multicolumn{6}{c}{\textbf{Supercomputer logs}} \\ \hline 
BGL & Blue Gene/L supercomputer log & 214.7 days & 708.76MB & 4,747,963 & \cmark & 348,460 & 619 \\ \hline 
HPC & High performance cluster log & N.A. & 32MB & 433,489 & \xmark & & 104\\ \hline 
Thunderbird & Thunderbird supercomputer log & 244 days & 29.6GB & 211,212,192 & \cmark & 3,248,239 & 4040 \\ \hline 
\multicolumn{6}{c}{\textbf{Operating System logs}} \\ \hline 
Windows & Windows event log & 226.7 days & 16.09GB & 114,608,388 & \xmark & & 4833\\ \hline 
Linux & Linux system log & 263.9 days & 2.25MB & 25,567 & \xmark & & 488\\ \hline 
Mac & Mac OS log & 7 days & 16.09MB & 117,283 & \xmark & & 2214 \\ \hline 
\multicolumn{6}{c}{\textbf{Mobile System logs}} \\ \hline 
Android & Android framework log & N.A. & 183.37MB & 1,555,005 & \xmark & & 76,923 \\ \hline 
Health App & Health app log & 10.5days & 22.44MB & 253,395 & \xmark & & 220\\ \hline 
\multicolumn{6}{c}{\textbf{Server application logs}} \\ \hline 
Apache & Apache server error logs & 263.9 days & 4.9MB & 56,481 & \xmark & & 44 \\ \hline
OpenSSH & OpenSSH server logs & 28.4 days & 70.02MB & 655,146 & \xmark & & 62 \\ \hline
\multicolumn{6}{c}{\textbf{Standalone software logs}} \\ \hline 
Proxifier & Proxifier software logs & N.A. & 2.42MB & 21,329 & \xmark & & 9 \\ \hline 
\multicolumn{6}{c}{\textbf{Hardware logs}} \\ \hline
Switch & Switch hardware failures & 2 years & - & 29,174,680 & \cmark & 2,204 & -\\ \hline
\end{tabular}
}
\end{table*}
\label{tab:log_datasets}

%% file: tables/log_ad_table.tex
\begin{table*}[hp]
\centering
\caption{Comparison of existing Log Anomaly Detection Models}
\begin{tabular}{|p{0.09\textwidth}|p{0.09\textwidth}|p{0.2\textwidth}|p{0.25\textwidth}|p{0.1\textwidth}|p{0.05\textwidth}|p{0.065\textwidth}|} \hline
 \textbf{Reference} & \textbf{Learning Setting} & \textbf{Type of Model} & \textbf{Log Representation} & \textbf{Log Tokens} & \textbf{Parsing} & \textbf{Sequence modeling}\\
 \hline
\cite{10.1145/1755913.1755926,4470294,1301345} & Supervised & Linear Regression, SVM, Decision Tree & handcrafted feature & log template &  \cmark & \xmark \\ \hline
\cite{DBLP:conf/sosp/XuHFPJ09} & Unsupervised & Principal Component Analysis (PCA) & quantitative & log template & \cmark & \cmark \\ \hline
\cite{DBLP:conf/cnsm/VaarandiP15,10.1145/2889160.2889232,10.1145/3236024.3236083,7381796} & Unsupervised &  Clustering and Correlation between logs and metrics & sequential, quantitative & log template & \cmark & \xmark \\ \hline
\cite{DBLP:conf/usenix/LouFYXL10} & Unsupervised & Mining invariants using singular value decomposition & quantitative, sequential & log template & \cmark & \xmark \\ \hline
\cite{10.1145/2939672.2939712,8029741,8030620,5360240} & Unsupervised & Frequent pattern mining from Execution Flow and control flow graph mining & quantitative, sequential & log template & \cmark & \xmark \\ \hline
\cite{10.1145/3468264.3473933,10.1145/3468264.3473919} & Unsupervised & Rule Engine over Ensembles and Heuristic contrast analysis over anomaly characteristics & sequential (with tf-idf weights) & log template & \cmark & \xmark \\ \hline
\cite{9209707} & Supervised & Autoencoder for log specific word2vec & semantic (trainable embedding) & log template & \cmark & \cmark \\ \hline
\cite{FARZAD2020229} & Unsupervised & Autoencoder w/ Isolation Forest & semantic (trainable embedding) & all tokens & \xmark & \xmark \\\hline
\cite{DBLP:conf/dasc/LuWLW18} & Supervised & Convolutional Neural Network & semantic (trainable embedding) & log template & \cmark & \cmark \\\hline
\cite{10.1145/3133956.3134015} & Unsupervised & \multirow{2}{*}{Attention based LSTM} & sequential, quantitative, semantic (GloVe embedding) & log template, log parameter & \cmark & \cmark \\ \hline
\cite{10.5555/3367471.3367702} & Unsupervised & Attention based LSTM & quantitative and semantic (GloVe embedding) & log template & \cmark & \cmark \\ \hline
\cite{10.1145/3338906.3338931} & Supervised & Attention based LSTM & semantic (fastText embedding with tf-idf weights) & log template & \cmark & \cmark \\ \hline
\cite{9401970} & Semi-Supervised &  Attention based GRU with clustering & semantic (fastText embedding with tf-idf weights) & log template & \cmark & \cmark \\ \hline
\cite{10.1145/3217871.3217872} & Unsupervised & Attention based Bi-LSTM & semantic (with trainable embedding) & all tokens & \xmark & \cmark \\ \hline
\cite{9527018} & Unsupervised & Bi-LSTM & semantic (token embedding from BERT, GPT, XLM) & all tokens & \xmark & \cmark \\ \hline
\cite{9251078} & Unsupervised & Attention based Bi-LSTM & semantic (BERT token embedding) & log template & \cmark & \cmark \\ \hline
\cite{9251092} & Semi-Supervised & LSTM, trained with supervision from source systems & semantic (GloVe embedding) & log template & \cmark & \cmark \\ \hline
\cite{10.1145/3459637.3482209} & Unsupervised & LSTM with domain adversarial training & semantic (GloVe embedding) & all tokens & \xmark & \cmark \\ \hline
\cite{https://doi.org/10.48550/arxiv.2210.04145,10.1145/3459637.3482209} & Unsupervised & LSTM with Deep Support Vector Data Description & semantic (trainable embedding) &  log template & \cmark & \cmark \\ \hline 
\cite{https://doi.org/10.48550/arxiv.2209.07869} & Supervised & Graph Neural Network & semantic (BERT token embedding) & log template & \cmark & \cmark \\ \hline
\cite{10.1007/978-3-030-75762-5_6} &  Semi-Supervised & Graph Neural Network & semantic (BERT token embedding) & log template & \cmark & \cmark \\ \hline
\cite{9338283,DBLP:journals/corr/abs-2101-02392,10.1145/3534678.3539155,9606230} & Unsupervised & Self-Attention Transformer & semantic (trainable embedding) & all tokens & \xmark & \cmark \\\hline
\cite{9678773} & Supervised & Self-Attention Transformer & semantic ( trainable embedding) & all tokens & \xmark & \cmark \\\hline
\cite{9244088} & Supervised & Hierarchical Transformer & semantic (trainable GloVe embedding) & log template, log parameter & \cmark & \cmark \\\hline
\cite{9401970,9534113} & Unsupervised & BERT Language Model & semantic (BERT token embedding) & all tokens & \xmark & \cmark \\\hline
\cite{DBLP:journals/corr/abs-2112-03159} & Unsupervised & Unified BERT on various log analysis tasks & semantic (BERT token embedding) & all tokens & \xmark & \cmark \\\hline
\cite{2021Entrp2469W} & Unsupervised & Contrastive Adversarial model & semantic (BERT and VAE based embedding) and quantitative & log template & \cmark & \cmark \\\hline 
\cite{10.1007/s10796-020-10026-3,10.1007/s10489-021-02863-9,DBLP:conf/noms/QiLHWFYQ22} & Unsupervised & LSTM,Transformer based GAN (Generative Adversarial) & semantic (trainable embedding) & log template & \cmark & \cmark \\\hline
\multicolumn{7}{l}{\textbf{Log Tokens} refers to the tokens from the logline used in the log representations}\\
\multicolumn{7}{l}{\textbf{Parsing} and \textbf{Sequence Modeling} columns respectively refers to whether these models need log parsing and they support modeling log sequences}\\

\end{tabular}
\label{tab:log_ad_comp}
\end{table*}

%% file: tables/metric_ad.tex
\begin{table*}[hp]
\centering
\caption{Comparison of Existing Metric Anomaly Detection Models}
\begin{tabular}{|p{0.1\textwidth}|p{0.15\textwidth}|p{0.2\textwidth}|p{0.1\textwidth}|p{0.1\textwidth}|p{0.15\textwidth}|}
\hline
\textbf{Reference}  & \textbf{Label Accessibility} & \textbf{Machine Learning Model} & \textbf{Dimensionality} & \textbf{Infrastructure} & \textbf{Streaming Updates} \\
\hline
\cite{liu2015opprentice} & Supervised & Tree & Univariate & \xmark    & \cmark (Retraining) \\
\hline
\cite{laptev2015generic} & Active & -     & Univariate & \cmark   & \cmark (Retraining) \\
\hline
\cite{guha2016robust} & Unsupervised & Tree & Multivariate & \xmark    & \cmark \\
\hline
\cite{ahmad2017unsupervised} & Unsupervised & Statistical & Univariate & \xmark    & \cmark \\
\hline
\cite{hochenbaum2017automatic} & Unsupervised & Statistical & Univariate & \xmark    & \xmark \\
\hline
\cite{bu2018rapid} & Semi-supervised & Tree & Univariate & \xmark    & \cmark \\
\hline
\cite{xu2018unsupervised} & Unsupervised, Semi-supervised & Deep Learning   & Univariate & \xmark    & \xmark \\
\hline
\cite{ren2019time} & Unsupervised & Deep Learning  & Univariate & \cmark   & \xmark \\
\hline
\cite{zhang2019cross} & Domain Adaptation, Active & Tree & Univariate & \xmark    & \xmark \\
\hline
\cite{su2019robust} & Unsupervised & Deep Learning & Multivariate & \xmark    & \xmark \\
\hline
\cite{ayed2020anomaly} & Unsupervised & Deep Learning & Univariate & \xmark    & \xmark \\
\hline
\cite{audibert2020usad} & Unsupervised & Deep Learning & Multivariate & \xmark    & \xmark \\
\hline
\cite{gao2020robusttad} & Supervised & Deep Learning & Univariate & \cmark   & \cmark (Retraining) \\
\hline
\cite{li2021multivariate} & Unsupervised & Deep Learning & Multivariate & \xmark    & \xmark \\
\hline
\cite{yang2022causality} & Unsupervised & Deep Learning & Multivariate & \xmark    & \xmark \\
\hline
\cite{rabanser2022intrinsic} & Unsupervised & Deep Learning & Multivariate & \xmark    & \xmark \\
\hline
\cite{huang2022semi} & Semi-supervised, Active & Deep Learning   & Multivariate & \cmark   & \cmark (Retraining) \\
\hline
\end{tabular}%
  \label{tab:metric_ad}%
\end{table*}%

%% file: tables/trace_ad_table.tex
\begin{table*}[hp]
  \centering
  \caption{Comparison of Existing Trace and Multimodal Anomaly Detection and RCA Models}
  \begin{tabular}{|p{0.2\textwidth}|p{0.25\textwidth}|p{0.2\textwidth}|p{0.2\textwidth}|}
  \hline
    \textbf{Reference} & \textbf{Topic} & \textbf{Deep Learning Adoption} & \textbf{Method}  \\\hline
    \cite{Chen2002} & Trace RCA & \xmark & Clustering \\\hline
    \cite{Arnold2007} & Trace RCA & \xmark & Heuristic \\\hline
    \cite{Attariyan2012} & Trace RCA & \xmark & Multi-input Differential Summarization \\\hline
    \cite{Zhou2019} & Trace RCA & \xmark & Random forest, k-NN \\\hline
    \cite{cai2019real} & Trace RCA & \xmark & Heuristic \\\hline
    \cite{Guo2020} & Trace Anomaly Detection & \xmark & Graph model \\\hline
    \cite{Liu2020} & Multimodal Anomaly Detection & \cmark & Deep Bayesian Networks \\\hline
    \cite{Xu2021} & Trace Representation & \cmark & Tree-based RNN \\\hline
    \cite{Li2021} & Trace Anomaly Detection & \xmark & Heuristic \\\hline
    \cite{Zhang2022} & Multimodal Anomaly Detection & \cmark & GGNN and SVDD \\\hline
    
  \end{tabular}
  \label{tab:trace_ad}
  \end{table*}

%% file: tables/metric_rca_table.tex
\begin{table*}[hp]
\centering
\caption{Comparison of several existing metric RCA approaches}
\begin{tabular}{|p{0.1\textwidth}|p{0.4\textwidth}|p{0.375\textwidth}|} \hline
 Reference & Metric or Graph Analysis & Root Cause Score  \\ \hline
 \cite{6681572} & Change points & Chronological order \\ \hline
 \cite{10.1145/2038633.2038634} & Change points & Chronological order \\ \hline
 \cite{10.1145/3308558.3313653} & Two-sample test & Correlation \\ \hline
 \cite{10.1145/3135974.3135977} & Call graphs & Cluster similarity \\ \hline
 \cite{9110353} & Service graph & PageRank \\ \hline
 \cite{BRANDON2020110432} & Service graph & Graph similarity \\ \hline
 \cite{8972825} & Service graph & Hierarchical HMM \\ \hline
 \cite{8411065} & PC algorithm & Random walk \\ \hline
 \cite{8367054} & ITOA-PI & PageRank \\ \hline
 \cite{7563819} & Service graph and PC & Causal inference \\ \hline
 \cite{9213058} & PC algorithm & Random walk \\ \hline
 \cite{Lin2018MicroscopePP} & Service graph and PC & Causal inference \\ \hline
 \cite{8818432} & PC algorithm & Random walk \\ \hline
 \cite{10.1145/3366423.3380111} & PC algorithm & Random walk \\ \hline
 \cite{10.1145/3534678.3539041} & Service graph & Causal inference \\ \hline
 \cite{Budhathoki2022} & Service graph & Contribution-based \\ \hline
\end{tabular}

\label{table:metric_rca}
\end{table*}